\documentclass[10pt,twocolumn,letterpaper]{article}
\usepackage[pagenumbers]{iccv} 

\usepackage{array}
\newcolumntype{x}[1]{>{\centering\arraybackslash}p{#1}}
\newcolumntype{y}[1]{>{\raggedright\arraybackslash}p{#1}}
\newcolumntype{z}[1]{>{\raggedleft\arraybackslash}p{#1}}
\newcommand{\tablestyle}[2]{\setlength{\tabcolsep}{#1}\renewcommand{\arraystretch}{#2}\centering}
\newcommand{\tablestylefontsize}[2]{\setlength{\tabcolsep}{#1}\renewcommand{\arraystretch}{#2}\centering\footnotesize}
\usepackage{makecell}

\usepackage{pifont}
\usepackage[table]{xcolor}
\usepackage{bm}
\usepackage{CJKutf8}
\usepackage{comment}
\usepackage{makecell}
\usepackage{utfsym}

\definecolor{iccvblue}{rgb}{0.21,0.49,0.74}
\usepackage[pagebackref,breaklinks,colorlinks,allcolors=iccvblue]{hyperref}

\usepackage{multirow}
\usepackage[hypcap=false]{caption}

\def\paperID{xxxx}
\def\confName{ICCV}
\def\confYear{2025}

\title{GPT4Scene: Understand 3D Scenes from Videos with Vision-Language Models}

\author{
Zhangyang Qi$^{1,2}$\footnotemark[1] \quad
Zhixiong Zhang$^{2}$\footnotemark[1] \quad
Ye Fang$^{2}$  \quad
Jiaqi Wang$^{2}$\footnotemark[2] \quad
Hengshuang Zhao$^{1}$\footnotemark[2] \\
$^{1}$The University of Hong Kong \quad $^{2}$Shanghai Artificial Intelligence Laboratory \\
{\tt\small \{zyqi, hszhao\}@cs.hku.hk, \{zhangzhixiong, fangye, wangjiaqi\}@pjlab.org.cn}\\
{\tt\small \url{https://gpt4scene.github.io}} \quad {\small * equal contribution} \quad {\small $\dag$ corresponding author}
}

\begin{document}

\twocolumn[{%
\renewcommand\twocolumn[1][]{#1}%
\maketitle
\begin{center}
\centering
\vspace{-5mm}
\includegraphics[width=0.98\textwidth]{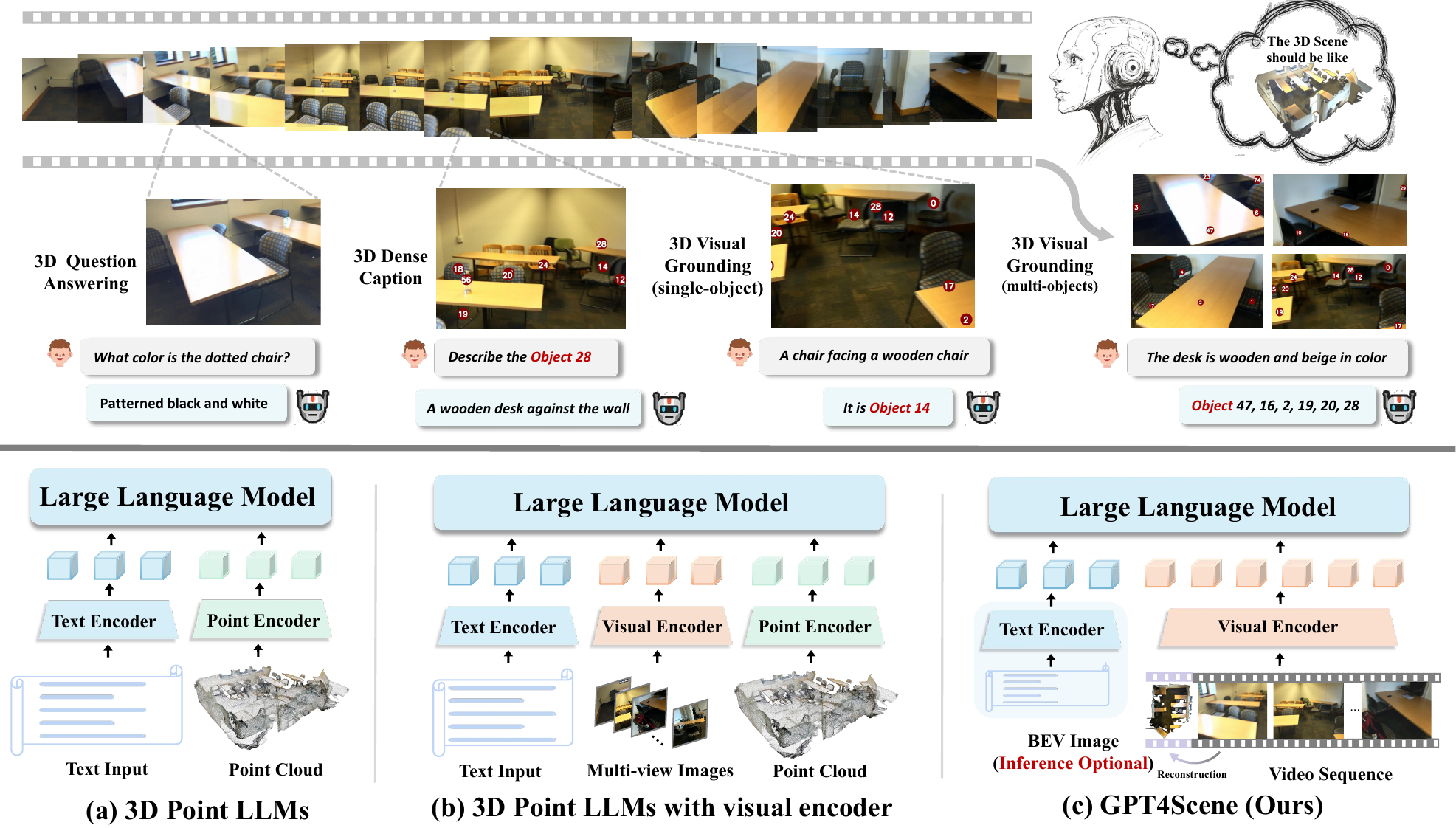}
\captionof{figure}
{
\textbf{The overall architecture of GPT4Scene.} It is capable of understanding 3D scenes and performing tasks such as 3D question answering, dense captioning, and visual grounding using only video input. In contrast to 3D point LLMs, GPT4Scene takes input solely from the vision modality, with global information provided by the BEV image reconstructed from the 3D structure derived from the video.
}
\label{fig_teaser}
\vspace{-1mm}
\end{center}
}] 

\begin{abstract}
In recent years, 2D Vision-Language Models (VLMs) have made significant strides in image-text understanding tasks. However, their performance in 3D spatial comprehension, which is critical for embodied intelligence, remains limited. Recent advances have leveraged 3D point clouds and multi-view images as inputs, yielding promising results. However, we propose exploring a purely vision-based solution inspired by human perception, which merely relies on visual cues for 3D spatial understanding. This paper empirically investigates the limitations of VLMs in 3D spatial knowledge, revealing that their primary shortcoming lies in the lack of global-local correspondence between the scene and individual frames.
To address this, we introduce GPT4Scene, a novel visual prompting paradigm in VLM training and inference that helps build the global-local relationship, significantly improving the 3D spatial understanding of indoor scenes. Specifically, GPT4Scene constructs a Bird's Eye View (BEV) image from the video and marks consistent object IDs across both frames and the BEV image. The model then inputs the concatenated BEV image and video frames with markers. In zero-shot evaluations, GPT4Scene improves performance over closed-source VLMs like GPT-4o.
Additionally, we prepare a processed video dataset consisting of 165K text annotation to fine-tune open-source VLMs, achieving state-of-the-art performance on all 3D understanding tasks. Surprisingly, after training with the GPT4Scene paradigm, VLMs consistently improve during inference, even without object marker prompting and BEV image as explicit correspondence. It demonstrates that the proposed paradigm helps VLMs develop an intrinsic ability to understand 3D scenes, which paves the way for a seamless approach to extending pre-trained VLMs for 3D scene understanding.
\end{abstract}

\section{Introduction}
\label{sec_introduction}
3D scene understanding aims to comprehend the overall layout of the surrounding complex environments and the spatial relationships between objects~\cite{scanqa, Scan2cap, Scanrefer}. It plays a crucial role in applications such as embodied intelligence, virtual reality, and smart cities~\cite{LEO, ll3da, HourVideo}. With the rapid development of Large Language Models~\cite{llama3, gemma2, gpt4, qwen25, internlm2}, Vision-Language Models have demonstrated impressive performance in image and video understanding~\cite{llava, llava1.5, LLaVA-OneVision, minigpt4, Qwen2VL}. Researchers have extended this paradigm to 3D perception by incorporating point clouds, aiming to improve scene understanding~\cite{3dllm, ll3da, Chat-3D, scenellm, Lexicon3D, LEO, Chat-scene, Robin3D}.

Recent 3D point LLMs leverage point clouds aligned with LLM features for indoor scene understanding~\cite{ll3da, 3dllm}. While combining point clouds and images~\cite{PQ3D, ll3da, Chat-scene} in Point-Vision-LLM paradigms improves performance through richer visual cues, aligning these modalities with text remains challenging. This complexity motivates exploration of vision-only solutions, inspired by humans’ natural ability to perceive 3D environments without point clouds, offering a promising direction for efficient scene analysis.

In this work, we aim to leverage pre-trained VLMs without modifying their architecture, maximizing their vision perception capabilities. However, their effectiveness in understanding immersive 3D indoor scenes remains limited. Our analysis shows that directly inputting scene videos into VLMs fails in 3D scene understanding due to two factors: \romannumeral1) the lack of a global scene representation, \romannumeral2) misalignment between per-frame local observations and their spatial-temporal context.

To address this, we propose GPT4Scene, a framework that enhances VLMs' spatial understanding (see~\Cref{fig_teaser}). We first perform 3D reconstruction on input videos to generate a Bird's Eye View (BEV) image, offering a comprehensive scene layout. Additionally, we introduce Spatial-Temporal Object markers (STO-markers) in both the BEV image and 2D frames. These markers maintain consistent object IDs across frames (temporal level) and align with the BEV image (spatial level), bridging the global-local relationship. Empirical results show that GPT4Scene remains robust to reconstruction quality and marker accuracy, as it prioritizes learning global-local correspondences over precise geometric reconstructions.

We first explored the effectiveness of GPT4Scene under a training-free approach. Experimental results revealed that it was notably effective for powerful large-scale VLMs such as Qwen2-VL-72B~\cite{Qwen2VL}, as well as closed-source models like GPT-4o~\cite{gpt4o} and Gemini-1.5-Pro~\cite{Gemini}, significantly enhancing their 3D scene understanding capabilities, even reaching levels comparable to previous state-of-the-art point-based methods. However, the improvements were limited on smaller-scale models. For smaller open-source vision-language models (VLMs), we introduce ScanAlign, a multimodal dataset comprising 165K aligned data pairs featuring STO-marker-annotated video frames, BEV images, and textual descriptions. Fine-tuning Qwen2-VL-7B on ScanAlign achieves state-of-the-art performance, delivering a \textbf{48\%} relative improvement (40.7 → 60.7 EM-1 score) on 3D question answering (SQA3D), surpassing the previous SOTA Chat-Scene by \textbf{11.0\%} (54.6 vs 60.7). The model demonstrates even stronger 3D visual grounding capabilities on Multi3DRef, outperforming Chat-Scene by \textbf{13.0\%} (57.1 → 64.5). These advancements underscore GPT4Scene's effectiveness in seamlessly enhancing VLMs with strong 3D spatial understanding.

Our paper makes these major contributions:
\begin{itemize}
\item We introduce GPT4Scene, a framework that enhances Vision-Language Models (VLMs) to comprehend 3D scenes directly from pure vision input.
\item We introduce two techniques: \romannumeral1) A 3D BEV image with global context information, and \romannumeral2) Spatial-Temporal Object markers (STO-markers) for spatial and temporal consistency across BEV image and video frames.
\item We introduce ScanAlign, a dataset comprising video frames, BEV images with STO-markers, and text annotations. Fine-tuning VLMs on this dataset significantly improves 3D understanding, even with raw video inputs.
\item GPT4Scene demonstrates robust performance in zero-shot and fine-tuning settings, achieving SOTA results across diverse 3D scene understanding tasks.
\end{itemize}

\section{Related Work}
\label{sec_related_work}
\noindent \textbf{3D Indoor Scene Understanding. }
3D indoor scene understanding allows robots to identify object positions, structures, and relationships within indoor environments, enabling question-and-answer interactions about the scene's content. This process combines 3D perception with large language models (LLMs). 3D perception, as a foundational component, is typically trained on common indoor datasets~\cite{ScanNet, Arkitscenes, Matterport3d, MultiScan, HM3D, Rio, Replica, ProcTHOR, hssd-200, Structured3d} using point clouds as input, supporting tasks like 3D object detection and instance segmentation~\cite{VoteNet, Pointgroup, 3DETR, Mask3D, SoftGroup, PTV3, SAM3D}. Recent advancements in 3D Vision-Language Learning (3D-VL) combine 3D perception tasks with natural language, introducing diverse textual annotations on datasets like ScanNet to support tasks such as 3D Question Answering~\cite{scanqa, sqa3d, 3dqa, 3DMV-VQA}, 3D Dense Captioning~\cite{Scan2cap}, and 3D Visual Grounding~\cite{Scanrefer, Multi3DRefer, Referit3d, Scanents3d, Intent3D}. Initial studies focus on single 3D-VL tasks~\cite{X-trans2cap, vote2cap-detr, EDA, LAR, Transrefer3d, MVT-3DVG, BUTD, 3D-SPS, Sat, 3DVG-Transformer, viewrefer}, while recent research introduces unified models for multiple tasks~\cite{3djcg, D3Net}. 2D vision-language pretraining (2D-VLP) has driven progress in 3D visual-language learning (3D-VL)~\cite{Semantic-abstraction, Clip-goes-3d, Openscene, OpenMask3D, ULIP, Pointclip, I2P-MAE, VLM-Grounder}, with recent 3D-VLP~\cite{3D-VisTA, PQ3D, 3dvlp, PLA, Regionplc, SceneVerse, EmbodiedScan} methods demonstrating that combining 2D visual cues with 3D point clouds enables complementary cross-modal alignment. This multimodal approach helps overcome geometric complexity and sparse annotation challenges in pure point cloud processing.

\noindent \textbf{3D Point Cloud LLMs. }
3D vision-language tasks aim to integrate 3D scene understanding with natural language processing. However, we aspire to go further by incorporating 3D content into large language models (LLMs) to achieve more natural human-computer interaction. Initially, this began with 3D point cloud LLMs. 3D point cloud LLMs take point clouds as input, enabling natural language generation and interaction in 3D scenes. Early 3D LLMs focused on object-level geometry and appearance~\cite{Point-Bind, GPT4Point, pointllm, shapellm}. Later, they expanded to indoor scenes, emphasizing spatial relationships among objects and overall scene features, often utilizing scene point clouds augmented with auxiliary 2D multi-view images~\cite{3dllm, ll3da, Chat-3D, scenellm, Lexicon3D}. To better capture object relationships, recent 3D LLMs decouple scene objects before feeding them into LLMs~\cite{LEO, Chat-scene}. Some approaches rely more heavily on visual inputs to determine scene context~\cite{HourVideo, LLaVA-3D, cc}. Here, we aim to explore whether pure visual inputs can better handle indoor scene understanding.

\noindent \textbf{Vision Language Models (VLMs). }
Vision Language Models (VLMs) are multimodal models that integrate visual and language processing capabilities, enabling the understanding and generation of combined image-text information. The origin of VLMs can be traced back to CLIP's 2D image-text pair pretraining~\cite{CLIP, CC12M, Laion-5b}, which laid the foundation for incorporating LLMs. Early VLMs used attention mechanisms or Q-Former to fuse image and text modalities before inputting them to LLMs~\cite{Blip, blip2, InstructBLIP, Flamingo, kosmos}. Later, an approach emerged that directly projects image features into the LLM using an MLP~\cite{llava, llava1.5, LLaVA-OneVision, minigpt4, Qwen2VL}, achieving better performance. Building on this, VLMs expanded into visual grounding tasks~\cite{cogvlm, kosmos2, Glamm, Lisa, internlmxcomposer2, Emu, Minigpt-v2} and further to video understanding by using spatiotemporal compression to process information from long image sequences~\cite{Video-llama, Video-llama2, Video-chatgpt, Valley, Minigpt4-video, LLaMA-VID, Moviechat, TimeChat, LLaVA-NeXT}. Currently, while some studies employ VLMs for indoor scene understanding, most remain either benchmark-oriented~\cite{vsibench, MVoT} or still rely on 3D-based methods~\cite{Agent3D-Zero, Video-3D-LLM}. This fundamentally reveals that VLMs cannot directly comprehend 3D scenes, making our core mission to empower them with 3D world interpretation capabilities.

\section{Methodology}
\label{sec_methodology}
\begin{figure*}[!t]
  \vspace{-3mm}
  \centering
  \includegraphics[width=\textwidth]{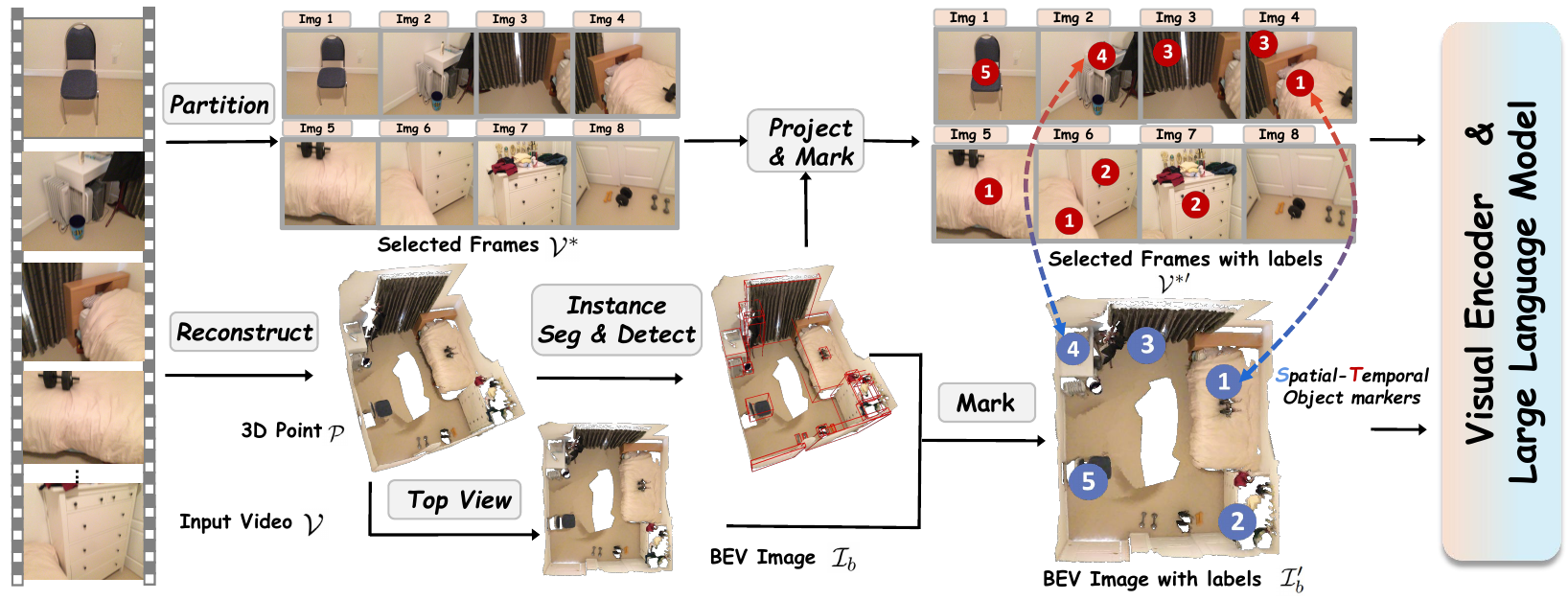}
  \caption{\textbf{The Framework of GPT4Scene. }A scene video is processed by sampling frames, reconstructing a point cloud, and generating a BEV image. Object locations are detected from the point cloud and projected onto the video frames. The resulting frames and BEV image, enhanced with STO-markers, are inputs for Large Language Model (VLM) training and inference.}
  \label{fig_training_process}
  \vspace{-1mm}
\end{figure*}

In this section, we introduce the GPT4Scene framework as shown in~\Cref{sec_gpt4cene_framework}, which enhances VLMs' 3D scene understanding using video inputs. In \Cref{sec_unlocking}, we discuss how we leverage zero-shot prompts to unlock the potential of powerful large-scale VLMs. In \Cref{sec_scanalign}, we apply fine-tuning to enhance VLMs for 3D understanding.

\subsection{GPT4Scene Framework}
\label{sec_gpt4cene_framework}
This section first presents an overview of the GPT4Scene framework, then details the 3D BEV Map (providing global scene layouts) and STO-markers (delivering localized object cues) as core components.

\vspace{2mm}
\noindent \textbf{Overview.}
Here we introduce the architecture of GPT4Scene. \Cref{fig_training_process} shows how GPT4Scene integrates global scene layouts and object-level details to improve VLMs' 3D scene understanding.
Given a video $\mathcal{V}\!=\!\{I_1, \ldots, I_N\}$ captured by moving around an indoor scene, we first approximately uniformly sample $n$ frames using indices:
$s_i = \left\lfloor (i-1)\frac{N}{n} \right\rfloor + 1, \quad \forall i \in \{1,\ldots,n\}$ to form $\mathcal{V}^*\!=\!\{I_{s_1}, \ldots, I_{s_n}\}$. This strategy reduces token counts and computational overhead for Vision-Language Models (VLMs) while preserving scene context, without significant information loss. We then leverage the complete temporal sequence for 3D scene reconstruction, generating a holistic Bird's-Eye View (BEV) map. Subsequent 3D instance segmentation enables precise object localization, which is projected onto both the BEV map and 2D video frames to establish Spatio-Temporal Object markers (STO-markers). The technical details are elaborated as follows.

\vspace{2mm}
\noindent \textbf{Global Information: 3D BEV Map.}
Egocentric videos lack global scene context. We address this by reconstructing the 3D scene into a point cloud and rendering a bird's-eye view (BEV) image for holistic VLM understanding. Given video \(\mathcal{V} = \{I_1, I_2, \ldots, I_N\}\) with camera extrinsics \(\mathcal{E} = \{E_1, E_2, \ldots, E_N\}\), we use 3D reconstruction techniques to generate 3D meshes and point clouds:
\begin{equation}
\mathcal{P} = \mathcal{R}\left( \{(I_t, E_t)\}_{t=1}^N \right)
\label{eq:reconstruction}
\end{equation}
Here, $\mathcal{R}(\cdot)$ denotes the reconstruction process, and we assume that the camera intrinsics are known. Then, we generate a BEV image of the scene from the global point cloud:
\begin{equation}
\mathcal{I}_{b} = \mathcal{T}(\mathcal{P}, \ E_{top})
\end{equation}
Here, $E_{\text{top}} \in \text{SE}(3)$ denotes the camera extrinsic for the top-down view, and \( \mathcal{T}(\cdot) \) represents the process of rendering the corresponding view based on the camera extrinsic, resulting in a BEV image of the scene. In particular, we continue to provide global 3D information to VLMs in the form of the top-view images instead of points.

\vspace{2mm}
\noindent \textbf{Local Correspondence: STO-markers.}
To help VLMs focus on specific objects, we introduce \textbf{\textcolor{iccvblue}{S}}patial-\textbf{\textcolor{red}{T}}emporal \textbf{O}bject markers (STO-markers), ensuring consistency between 2D frames and the 3D BEV image. Obtain the 3D point cloud \(\mathcal{P}\) reconstructed from the input video \(\mathcal{V}\). Applying 3D instance segmentation like Mask3D~\cite{Mask3D} yields instance masks \(\mathcal{M}=\{M_1, M_2, \ldots M_K\}\), where \(K\) denotes the total number of objects in the scene. 

\vspace{1mm}
For the BEV image $\mathcal{I}_{\bm{b}}$, we begin by projecting the 3D segmentation masks \(\mathcal{M}\) onto the \(xy\) plane. Then we extract the center coordinates of the resulting bounding box, denoted as 
$\bm{C}^{\bm{x}\bm{y}}=\{C^{xy}_{1}, C^{xy}_{2}, \ldots, C^{xy}_{K}\}$. 
These coordinates are subsequently overlaid onto the BEV image.

\vspace{1mm}
For selected egocentric video frames $\mathcal{V}^*$, the 3D instance masks $\mathcal{M}$ are projected onto each frame according to its corresponding camera pose. For each frame $i$, we extract the centroid of the 2D mask of each object as its 2D marker, defined as $\boldsymbol{C}^{\bm{u}\bm{v}}_{i} = \{ C^{uv}_{i,1}, C^{uv}_{i,2}, \ldots, C^{uv}_{i,K} \}$
, where $C^{uv}_{i,k}$ represents the 2D marker for the $k$-th object in the $i$-th frame. The processed 2D frames with markers and the BEV image with markers are defined as follows:
\begin{align}
\mathcal{V}^{*\prime} &= \left\{ \mathcal{F}\left(I_{i}, \ \boldsymbol{C}_{i}^{\bm{u}\bm{v}}\right) \mid i = s_1,s_2,\dots,s_n \right\} \\
\mathcal{I}_{b}^{\prime} &= \mathcal{F}\left(\mathcal{I}_{\bm{b}}, \ \boldsymbol{C}^{\bm{x}\bm{y}}\right)
\end{align}
Here, \( \mathcal{F}(\cdot) \) represents the operation of overlaying markers onto images, \( \mathcal{V}^{*\prime} \) and \( \mathcal{I}_{b}^{\prime} \) denote the video frames and BEV image with STO-markers. Notably, the 2D markers in each frame and their corresponding 3D markers in the BEV image maintain spatial alignment, ensuring that they represent the same objects. Moreover, \( \boldsymbol{C}^{\bm{u}\bm{v}}_{i} \) remains consistent across frames at the object level, preserving temporal coherence.

\subsection{Unlocking VLMs with Zero-shot Prompts}
\label{sec_unlocking}
~~~~We evaluate zero-shot VLMs, focusing on powerful closed-source models (e.g., GPT-4o) to test GPT4Scene’s effectiveness in 3D scene understanding. This process is termed ``unlocking'', which enables training-free 3D scene comprehension in VLMs via prompts. Specifically, we input $\mathcal{V}^{*\prime}$ and $\mathcal{I}_{b}^{\prime}$. To reduce the cost, we stitch the images from $\mathcal{V}^{*\prime}$ into a single large image. We evaluate three tasks: (1) In 3D question answering, the goal is to answer scene-related questions like, ``What is the color of the floor?". (2) In dense captioning, the task is to describe a specific object, such as ``Describe the object represented by $C_{5}$." (3) In visual grounding, the aim is to identify the object ID from a description, like ``What is the ID of the black chair next to the window?" While question answering is independent of object labels, dense captioning, and visual grounding require object markers. These tasks involve detecting objects and filtering based on the IoU of their bounding boxes. Consistent with Chat-Scene~\cite{Chat-scene} and Robin3D~\cite{Robin3D}, we use Mask3D segmentation results as predicted bounding boxes to calculate the IoU.

In addition to traditional tasks, we have conducted further experiments in this zero-shot setting. Qualitative results are shown in \Cref{fig_qualitative_results}. By inputting $\mathcal{V}^{*\prime}$ and $\mathcal{I}_{b}^{\prime}$, VLMs can understand the global features of indoor scenes. At this point, GPT-4o can still accept additional first-person perspective frames, allowing it to understand the current position in the scene to plan the following action. Additionally, using GPT-4o as an agent, VLMs can determine the task type based on the given question and choose the appropriate prompt. Therefore, the GPT4Scene framework shows excellent potential as a core technology for the next generation of embodied intelligence.

\begin{table}[t]
    \begin{minipage}{0.45\textwidth}
    \centering
        \tablestylefontsize{6pt}{1.10}
        \resizebox{\linewidth}{!}{
            \begin{tabular}{cc|c|cc}
\toprule
\multicolumn{2}{c|}{\makecell{\textbf{3D Question Answering} \\ (Scene Level)}} & 
\makecell{\textbf{3D Dense Caption} \\ (Object Level)} & 
\multicolumn{2}{c}{\makecell{\textbf{3D Visual Grounding} \\ (Object Level)}} \\
\midrule
ScanQA & SQA3D & Scan2cap & Multi3DRef & ScanRefer \\
26,138 & 26,623 & 35,056 & 41,408 & 35,061 \\
\midrule
\rowcolor{gray!10} \multicolumn{5}{c}{\textbf{Total Samples: 164,286}} \\
\bottomrule
\end{tabular}

        }
        \caption{\textbf{Text Annotations of ScanAlign.} We obtained the text annotations of ScanAlign by diversifying the text annotations related to ScanNet, resulting in 165K text annotations. }
        \label{tab_scanalign}
    \end{minipage}
\end{table}

\subsection{Enhancing VLMs with ScanAlign Fine-Tuning}
\label{sec_scanalign}
~~~~Zero-shot prompts can unlock the 3D understanding capabilities of powerful VLMs, but as shown in ~\Cref{tab_zero_shot_setting}, this approach does not improve smaller VLMs. Therefore, we aim to enhance open-source, smaller VLMs through fine-tuning. We first construct an indoor scene dataset, \textbf{ScanAlign}, with egocentric, BEV images and text annotations based on the ScanNet~\cite{ScanNet}. The dataset includes three 3D vision-related tasks represented as $(\mathcal{V}^{*\prime}, \mathcal{I}_{b}^{\prime}, T)$. The visual input consists of selected video frames with STO-markers and BEV images, and $T$ denotes text annotations derived from five ScanNet annotations~\cite{ScanNet}, as shown in \Cref{tab_scanalign}. We use prompts to randomly vary the annotation format to increase annotation diversity, with further details in the supplementary materials. The dataset contains 165K annotations in total. 

\vspace{1mm}
Since our method does not require additional modality alignment steps, we can directly perform single-stage instruction fine-tuning on the \textbf{ScanAlign} dataset to enhance the model's 3D spatial understanding capabilities. During the training phase, the training loss is the Cross-Entropy loss of the language model. The goal is to optimize the learnable parameters, denoted as $\theta$, by minimizing the negative log-likelihood of the target answer, $t^{a}$; we unify the system messages and the user's question as $t^{q}$. Therefore, the loss function can be expressed as follows:

\begin{equation}
\mathcal{L}(\theta)=-\sum_{i=1}^{k}\log P(t^{a}_{i}|t^{a}_{[1, \ldots, i-1]}, t^q),
\end{equation}
$k$ represents the number of tokens in the response sequence, and $t^{a}[1,..,i-1]$ denotes the previous $i-1$ tokens in the response. The set of learnable parameters $\theta$ is the vision language projection layers.

\vspace{1mm}
After fine-tuning with ScanAlign, we observed significant performance gains for smaller models, which proves that our approach effectively enhances the intrinsic 3D perception capabilities of visual language models (VLMs). We also conducted a series of ablation experiments to validate this finding.

\section{Experiments}
\label{sec_experiments}
\begin{table}[t]
    \hspace{-3mm}
    \begin{minipage}{0.48\textwidth}
    \centering
        \tablestyle{12pt}{1.1}
        \resizebox{\linewidth}{!}{
        \begin{tabular}{l|cc|cc}
\toprule
\multirow{2}{*}{\textbf{Zero-shot 3D QA}} & \multicolumn{2}{c}{ROUGE@\textbf{ScanQA}} & \multicolumn{2}{c}{EM-1@\textbf{SQA3D}} \\ 
\cmidrule(lr){2-5} 
 & VID & \small+GPT4Scene & VID & \small+GPT4Scene \\
\midrule

\multicolumn{5}{l}{\textcolor{gray}{\textit{Open-sourced VLM Based Model}}} \vspace{1mm} \\
InternVL2-8B~\cite{InternVL}  & 34.3 & 33.7$\,_\text{\textcolor{darkgray}{-0.6}}$ & 33.0 & 31.4$\,_\text{\textcolor{darkgray}{-1.6}}$  \\
MiniCPM-V-2.6-8B~\cite{MiniCPM-V}  &  31.5 & 32.1$\,_\text{\textcolor{darkgray}{+0.6}}$ & 42.6 & 43.3$\,_\text{\textcolor{darkgray}{+0.7}}$  \\
Qwen2-VL-2B~\cite{Qwen2VL} & 28.2 & 28.4$\,_\text{\textcolor{darkgray}{+0.2}}$ & 35.7 & 34.8$\,_\text{\textcolor{darkgray}{-0.9}}$ \\
Qwen2-VL-7B~\cite{Qwen2VL} & 29.3 & 31.7$\,_\text{\textcolor{darkgray}{+2.4}}$ & 40.7 & 41.7$\,_\text{\textcolor{darkgray}{+1.0}}$ \\
Qwen2-VL-72B~\cite{Qwen2VL} & 30.4 & 33.4$\,_\text{\textcolor{iccvblue}{\textbf{+3.0}}}$ & 39.8 & 42.3$\,_\text{\textcolor{iccvblue}{\textbf{+2.5}}}$ \\

\midrule
\multicolumn{5}{l}{\textcolor{gray}{\textit{Closed-sourced VLM Based Model}}} \vspace{1mm} \\
GPT-4o~\cite{gpt4o}  & 32.6 & 37.7$\,_\text{\textcolor{iccvblue}{\textbf{+5.1}}}$ & 40.3 & 42.8$\,_\text{\textcolor{iccvblue}{\textbf{+2.5}}}$  \\
Gemini-1.5-Pro~\cite{Gemini}  & 33.4 & 37.5$\,_\text{\textcolor{iccvblue}{\textbf{+4.1}}}$ & 41.7 & 44.2$\,_\text{\textcolor{iccvblue}{\textbf{+2.5}}}$  \\

\midrule
\multicolumn{1}{l}{\textcolor{gray}{\textit{3D LLM Based Model}}} & \multicolumn{2}{c}{\textcolor{gray}{\textit{Pre SOTA}}} & \multicolumn{2}{c}{\textcolor{gray}{\textit{Pre SOTA}}} \vspace{1mm} \\
Chat-Scene~\cite{Chat-scene} & \multicolumn{2}{c}{\textit{41.6}} & \multicolumn{2}{c}{\textit{54.6}}  \\

\bottomrule
\end{tabular}

        }
        \vspace{-2mm}
        \caption{\textbf{Zero-shot Setting Results.} GPT4Scene yielded limited enhancements for smaller models (2B, 7B, and 8B), while demonstrating significant improvements for larger architectures (72B, closed-source models). This disparity spurred further exploration of fine-tuning strategies for smaller-scale VLMs.}
        \label{tab_zero_shot_setting}
    \end{minipage}
    \vspace{-4mm}
\end{table}

\begin{table*}[!t]
    \begin{subtable}{\linewidth}
    \centering
    \vspace{2mm}
    \tablestyle{6pt}{1.0}
    \resizebox{\linewidth}{!}{
        \begin{tabular}{lccccccc| cc}
\toprule
3D Question Answering & \multirow{2}{*}{\makecell[c] {Point \\ Encoder}} & \multirow{2}{*}{\makecell[c] {Vision \\ Encoder}}& \multicolumn{5}{c}{\textbf{ScanQA (val)}} & \multicolumn{2}{c}{\textbf{SQA3D (val)}}
\\\cmidrule(lr){4-8}\cmidrule(lr){9-10}
Methods & & & BLEU-1 & BLEU-4 & METEOR & ROUGE & CIDEr & EM-1 & EM-R1 
\\
\midrule
\multicolumn{5}{l}{\textit{Task-Specific Model}} \vspace{1mm} \\
ScanQA~\cite{scanqa} & \textcolor{Green}{\ding{51}} & \textcolor{Red}{\ding{55}} & 30.2 & 10.1 & 13.1 & 33.3 & 64.9 & - & - \\
SQA3D~\cite{sqa3d} & \textcolor{Green}{\ding{51}} & \textcolor{Red}{\ding{55}} & - & - & - & - & - & 46.6 & - \\
3D-VLP~\cite{3dvlp} & \textcolor{Green}{\ding{51}} & \textcolor{Red}{\ding{55}} & 30.5 & 11.2 & 13.5 & 34.5 & - & - & - \\
3D-Vista~\cite{3D-VisTA} & \textcolor{Green}{\ding{51}} & \textcolor{Red}{\ding{55}} & - & - & 13.9 & 35.7 & - & 48.5 & - \\
\midrule
\multicolumn{5}{l}{\textit{3D LLM Based Model}} \vspace{1mm} \\
Chat-3D~\cite{Chat-3D} & \textcolor{Green}{\ding{51}} & \textcolor{Red}{\ding{55}} & 29.1 & 6.4 & 11.9 & 28.5 & 53.2 & - & - \\
Chat-3D v2~\cite{Chat-3Dv2} & \textcolor{Green}{\ding{51}} & \textcolor{Red}{\ding{55}} & 38.4 & 7.3 & 16.1 & 40.1 & 77.1 & - & - \\
3D-LLM~\cite{3dllm} & \textcolor{Green}{\ding{51}} & \textcolor{Green}{\ding{51}} & 39.3 & 12.0 & 14.5 & 35.7 & 69.4 & - & - \\
LL3DA~\cite{ll3da}  & \textcolor{Green}{\ding{51}} & \textcolor{Red}{\ding{55}} & - & 13.5 & 15.9 & 37.3 & 76.8 & - & - \\
PQ3D~\cite{PQ3D}   & \textcolor{Green}{\ding{51}} & \textcolor{Green}{\ding{51}} & - & - & - & - & - & 47.1 & - \\
LEO~\cite{LEO}  & \textcolor{Green}{\ding{51}} & \textcolor{Green}{\ding{51}} & - & 11.5 & 16.2 & 39.3 & 80.0 & 50.0 & 52.4 \\
\rowcolor{gray!10} Chat-Scene~\cite{Chat-scene} & \textcolor{Green}{\ding{51}} & \textcolor{Green}{\ding{51}} & 43.2 & 14.3 & {18.0} & 41.6 & 87.7 & 54.6 & 57.5 \\
\midrule

\multicolumn{5}{l}{\textit{Vision LLM Based Model}} \vspace{1mm} \\

\rowcolor{green!5} Qwen2-VL-7B~\cite{Qwen2VL}  & \textcolor{Red}{\ding{55}} & \textcolor{Green}{\ding{51}} & 27.8 & 3.0 & 11.4 & 29.3 & 53.9 & 40.7 & 46.7 \\

\rowcolor{green!7} Qwen2-VL-7B (GPT4Scene) & \textcolor{Red}{\ding{55}} & \textcolor{Green}{\ding{51}} & {43.4}$\,_\text{\textcolor{darkgray}{+15.6}}$ & {14.6}$\,_\text{\textcolor{darkgray}{+11.6}}$ & 17.7$\,_\text{\textcolor{darkgray}{+6.3}}$ & {43.6}$\,_\text{\textcolor{darkgray}{+14.2}}$ & {90.9}$\,_\text{\textcolor{darkgray}{+37.0}}$ & {57.4}$\,_\text{\textcolor{darkgray}{+16.7}}$ & {60.7}$\,_\text{\textcolor{darkgray}{+14.1}}$ \\

\rowcolor{green!10} Qwen2-VL-7B (GPT4Scene)-HD & \textcolor{Red}{\ding{55}} & \textcolor{Green}{\ding{51}} & {41.9}$\,_\text{\textcolor{darkgray}{+14.1}}$ & {\textbf{15.9}}$\,_\text{\textcolor{darkgray}{+12.9}}$ & 17.6$\,_\text{\textcolor{darkgray}{+6.2}}$ & {43.6}$\,_\text{\textcolor{darkgray}{+14.2}}$ & {89.9}$\,_\text{\textcolor{darkgray}{+36.0}}$ & {57.2}$\,_\text{\textcolor{darkgray}{+16.5}}$ & {60.4}$\,_\text{\textcolor{darkgray}{+13.5}}$ \\

\rowcolor{green!15} Qwen2-VL-7B (GPT4Scene)-HDM & \textcolor{Red}{\ding{55}} & \textcolor{Green}{\ding{51}} & {\textbf{44.4}$\,_\text{\textcolor{darkgray}{+16.6}}$} & {15.5$\,_\text{\textcolor{darkgray}{+12.5}}$} & \textbf{18.9}$\,_\text{\textcolor{darkgray}{+7.5}}$ & {\textbf{46.5}}$\,_\text{\textcolor{darkgray}{+17.1}}$ & {\textbf{96.3}}$\,_\text{\textcolor{darkgray}{+42.4}}$ & {\textbf{60.6}}$\,_\text{\textcolor{darkgray}{+19.9}}$ & {\textbf{63.3}}$\,_\text{\textcolor{darkgray}{+16.4}}$ \\

\bottomrule
\end{tabular}

    }
    \end{subtable}
    \vspace{2mm}
    \caption{\textbf{Evaluation of 3D Question Answering on ScanQA~\cite{scanqa} and SQA3D~\cite{sqa3d} datasets. }GPT-4o (GPT4Scene) in the zero-shot setting outperforms most 3D LLM models. Fine-tuned with GPT4Scene, Qwen2-VL achieves state-of-the-art performance. The \textbf{base} setting uses $N=8$ frames at $128 \times 123$, "\textbf{HD}" increases resolution to $512 \times 490$, and "\textbf{HDM}" combines this resolution with $N=32$ frames.}
    \label{tab_3d_qa}
    \vspace{-1mm}  
\end{table*}

~~~~In this section, we primarily present the experimental results. First, we evaluate the improvements of GPT4Scene under zero-shot settings in \Cref{sec_zero_shot_setting_results}. Following this, we outline the training details in \Cref{sec_implementation_details}, and subsequently introduce the fine-tuning results in \Cref{sec_fine_tuning_main_results}. Finally, \Cref{sec_ablation_study} details the ablation study, demonstrating the effectiveness of individual components.

\subsection{Zero-shot Setting Results}
\label{sec_zero_shot_setting_results}
\begin{flushright}
    \vspace{-0.5mm}
    \small\textcolor{gray}{\textit{Can VLMs understand the 3D world directly?}}
    \vspace{-2mm}
\end{flushright}
~~~~First, we examine GPT4Scene's zero-shot performance, with the results detailed in \Cref{tab_zero_shot_setting}. We begin by comparing a range of open-source models, including InternVL2-8B~\cite{InternVL}, MiniCPM-V-2.6-8B~\cite{MiniCPM-V}, and the Qwen2-VL family (2B, 7B, and 72B)~\cite{Qwen2VL}. Our analysis reveals that for smaller models (2B, 7B, and 8B), the integration of GPT4Scene demonstrates limited effectiveness in enhancing 3D spatial understanding through VLMs. This suggests that the benefits of GPT4Scene are more pronounced in larger models, highlighting the need for further optimization and architectural advancements to bridge this gap.

However, significant improvements are observed with larger models like Qwen2-VL-72B. When testing API-based open-source models Gemini-1.5-Pro~\cite{Gemini} and GPT-4o~\cite{gpt4o}, we find that incorporating BEV and STO-markers yields substantial gains, achieving performance comparable to previous state-of-the-art methods Chat-Scene~\cite{Chat-scene}. These results suggest that smaller VLMs face inherent limitations in visual comprehension capabilities, as they cannot directly enhance 3D scene understanding through zero-shot prompting alone. Consequently, we propose constructing a specialized dataset, ScanAlign, and conducting fine-tuning on Qwen2-VL to achieve performance improvements.

\subsection{Implementation Details}
\label{sec_implementation_details}
~~~~Our 3D scene understanding benchmark is built upon the ScanNet dataset (1,513 scenes, following its original data splits) and encompasses three core tasks: 3D question answering (evaluated via ScanQA~\cite{scanqa} and SQA3D~\cite{sqa3d}), 3D dense captioning (Scan2Cap~\cite{Scan2cap}), and 3D visual grounding (ScanRefer~\cite{Scanrefer} and Multi3DRef~\cite{Multi3DRefer}). The data processing pipeline includes reconstructing point clouds with BundleFusion~\cite{BundleFusion} (the original ScanNet method), performing 3D instance segmentation with Mask3D~\cite{Mask3D} (consistent with Chat-Scene~\cite{Chat-scene} and Robin3D~\cite{Robin3D}), projecting object centers to bird's-eye-view coordinates, and mapping 3D masks onto 2D images for marker positioning. 2D marker placement is necessary for dense captioning and visual grounding tasks. 

For the experienment config, we sample N=8 frames per video (128×123 resolution). Closed-source visual language models (VLMs) like GPT-4o process stitched 8-frame panoramas, while open-source models (e.g., Qwen2-VL-7B) directly take in sampled frames. The open-source models also support enhanced configurations: \textbf{HD} mode (high resolution, 512×490) and \textbf{HDM} mode (high resolution with multiple frames, 512×490, 32 frames). Training is conducted for one epoch using a base learning rate of 5e-6 with cosine annealing scheduling, completing in approximately 6 hours on 8×A100 GPUs. 2D marker localization is preserved to meet the specific needs of dense captioning and visual grounding tasks.

\subsection{Fine-tuning Main Results}
\label{sec_fine_tuning_main_results}
We present our experimental results using the ScanAlign dataset for VLM fine-tuning. The results demonstrate that using only Qwen2-VL-7B~\cite{Qwen2VL}, we achieve state-of-the-art performance across all indoor scene understanding tasks on ScanNet. This validates the effectiveness of GPT4Scene in enhancing VLMs' capacity for 3D scene comprehension.

\vspace{2mm}
\noindent \textbf{3D Question Answering. }
The 3D question answering results are presented in \Cref{tab_3d_qa}. We categorize the evaluated approaches into three groups: (1) classic task-specific models optimized for 3D QA tasks, (2) 3D point cloud LLM-based architectures, and (3) vision-language multimodal LLM systems. Our experimental analysis demonstrates that the baseline Qwen2-VL-7B model without fine-tuning shows constrained capability in 3D QA scenarios. Through systematic fine-tuning with GPT4Scene, the enhanced Qwen2-VL-7B(GPT4Scene) achieves state-of-the-art performance in 3D visual question answering, surpassing all comparative methods across evaluation metrics. Quantitative comparisons reveal 56.1\% improvement in BLEU-1 (27.8 → 43.4) and 68.6\% enhancement in CIDEr (53.9 → 90.9) on the ScanQA benchmark compared to the original Qwen2-VL-7B implementation. For SQA3D evaluations, we observe a 41.0\% increase in EM-1 scores (40.7 → 57.4). Furthermore, under HD (High-Resolution) and HDM (High-Resolution \& multi frames) configurations, the model exhibits additional performance gains.

\begin{table}[!t]
  \centering
  \hspace{-3mm}
  \begin{minipage}{0.47\textwidth}
    \vspace{-1.1mm}
    \tablestyle{4pt}{1.08}
    \resizebox{\linewidth}{!}{
      \begin{tabular}{lcccc}
\toprule
3D Dense Caption & \multicolumn{2}{c}{IoU@0.25} & \multicolumn{2}{c}{IoU@0.5} \\
\cmidrule(lr){2-3} \cmidrule(lr){4-5}
Methods & BLEU-4 & ROUGE & BLEU-4 & ROUGE \\

\midrule
\multicolumn{5}{l}{\textcolor{gray}{\textit{Task-Specific Model}}} \vspace{1mm} \\

Scan2Cap~\cite{Scan2cap} & 34.2 & 55.3 & 23.3 & 44.5 \\

3DJCG~\cite{3djcg} & 40.2 & 59.2 & 31.0 & 50.8 \\

X-Trans2Cap~\cite{X-trans2cap} & 35.7 & 54.7 & 25.1 & 45.3 \\

3D-VisTA~\cite{3D-VisTA} & 36.5 & 57.6 & 34.0 & 54.3 \\

Vote2Cap-DETR~\cite{vote2cap-detr} & 39.3 & 59.3 & 34.5 & 54.4 \\

\midrule
\multicolumn{5}{l}{\textcolor{gray}{\textit{3D LLM Based Model}}} \vspace{1mm} \\

LL3DA~\cite{ll3da} & 41.4 & 59.5 & 36.8 & 55.1 \\

LEO~\cite{LEO} & -- & -- & 36.9 & 57.8 \\

Chat-Scene~\cite{Chat-scene} & 38.2 & 60.6 & 36.3 & 58.1 \\

Robin3D~\cite{Robin3D} & -- & -- & 38.4 & -- \\

\midrule
\multicolumn{5}{l}{\textcolor{gray}{\textit{Vision LLM Based Model}}} \vspace{1mm} \\

\rowcolor{green!5} Qwen2-VL-7B~\cite{Qwen2VL} & 3.8 & 24.7 & 3.8 & 24.6\\

\rowcolor{green!7} Qwen2-VL-7B (GPT4Scene)  & 36.3$\,_\text{\textcolor{darkgray}{+32.5}}$  & 57.6$\,_\text{\textcolor{darkgray}{+32.9}}$ & 34.2$\,_\text{\textcolor{darkgray}{+30.4}}$ & 55.2$\,_\text{\textcolor{darkgray}{+30.6}}$ \\

\rowcolor{green!10} Qwen2-VL-7B (GPT4Scene)-HD  & 40.4$\,_\text{\textcolor{darkgray}{+36.6}}$  & 60.2$\,_\text{\textcolor{darkgray}{+35.5}}$ & 37.9$\,_\text{\textcolor{darkgray}{+34.1}}$ & 57.7$\,_\text{\textcolor{darkgray}{+33.1}}$ \\

\rowcolor{green!15} Qwen2-VL-7B (GPT4Scene)-HDM  & \textbf{43.1}$\,_\text{\textcolor{darkgray}{+39.3}}$  & \textbf{61.9}$\,_\text{\textcolor{darkgray}{+37.2}}$ & \textbf{40.6}$\,_\text{\textcolor{darkgray}{+36.8}}$ & \textbf{59.3}$\,_\text{\textcolor{darkgray}{+34.7}}$ \\

\bottomrule
\end{tabular}

    }
    \vspace{-2mm}
    \caption{\textbf{Evaluation of 3D Dense Caption on Scan2Cap~\cite{Scan2cap}.} Our results outperform those of existing 3D LLM based models.}
    \label{tab_3d_dense_caption}
  \end{minipage}
  \vspace{-4mm}
\end{table}

\begin{figure}[!t]
  \centering
  \begin{minipage}{0.47\textwidth}
    \vspace{-1mm}
    \includegraphics[width=\textwidth]{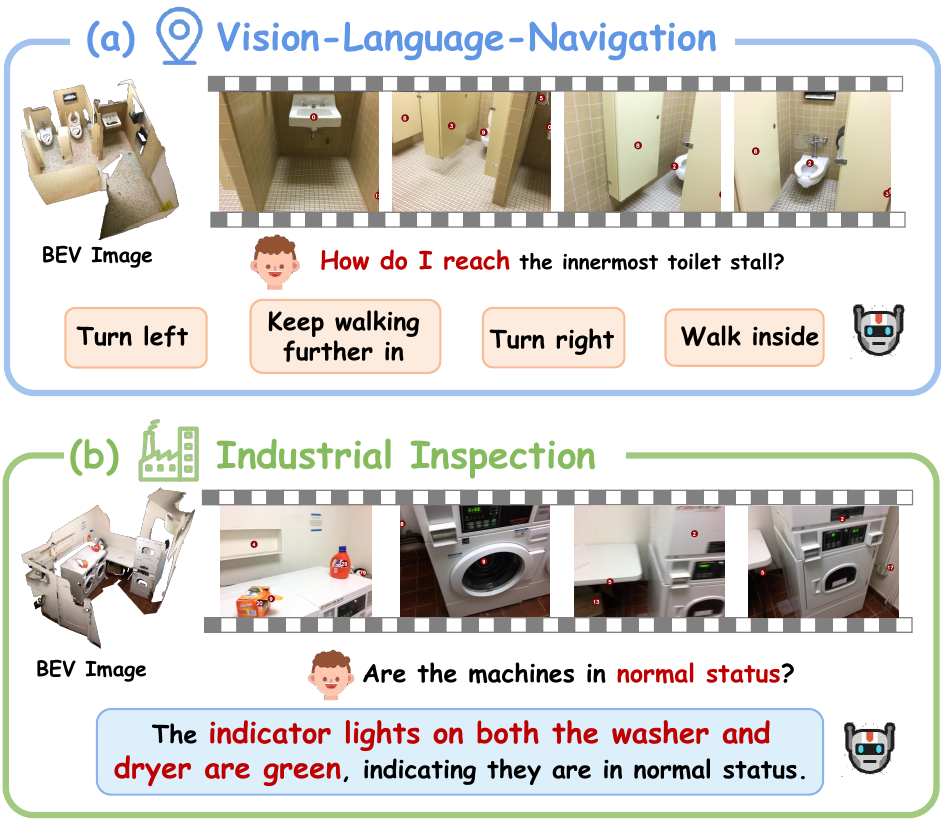}
    \vspace{-2mm}
    \caption{\textbf{Qualitive Results. }We demonstrate the qualitative performance of GPT-4o under zero-shot settings using GPT4Scene prompts, highlighting two distinctive application cases that exemplify the versatility of GPT4Scene.}
    \label{fig_qualitative_results}
  \end{minipage}
  \vspace{-4mm}
\end{figure}

\vspace{2mm}
\noindent \textbf{3D Dense Caption \& Visual Grounding. }
We further assess the model's capabilities in dense captioning and visual grounding tasks, which differ fundamentally from question answering by requiring explicit grounding mechanisms between textual descriptions and 3D spatial structures. As evidenced by the experimental results in \Cref{tab_3d_dense_caption} and \Cref{tab_visual_grounding}, our GPT4Scene-finetuned Qwen2-VL-7B demonstrates substantial gains in both 3D caption generation and object localization accuracy compared to baseline vision-language models (VLMs). The performance improvements become particularly pronounced under HD and HDM configurations, where our approach establishes new state-of-the-art results across all evaluation metrics, exceeding existing methods by significant margins.

\begin{table}[!t]
    \hspace{-3mm}
    \begin{minipage}{0.47\textwidth}
    \centering
        \vspace{-1.1mm}
        \tablestyle{4pt}{1.17}
        \resizebox{\linewidth}{!}{
        \begin{tabular}{lcccc}
\toprule
3D Visual Grounding & \multicolumn{2}{c}{ScanRefer} & \multicolumn{2}{c}{Multi3DRef} \\
\cmidrule(lr){2-3} \cmidrule(lr){4-5}
Methods & Acc@0.25 & Acc@0.50 & all F1@0.25 & all F1@0.50 \\
\midrule
\multicolumn{5}{l}{\textcolor{gray}{\textit{Task-Specific Model}}} \vspace{1mm} \\

3DVG-Transformer~\cite{3DVG-Transformer} & 47.6 & 34.7 & -- & 25.5 \\

3DJCG~\cite{3djcg} & 49.6 & 37.3 & -- & 26.6 \\

D3Net~\cite{D3Net}  & -- & 37.9 & -- & 32.2 \\

M3DRef-CLIP~\cite{Multi3DRefer} & 51.9 & 44.7 & 42.8 & 38.4 \\

\midrule
\multicolumn{5}{l}{\textcolor{gray}{\textit{3D LLM Based Model}}} \vspace{1mm} \\

Chat-Scene~\cite{Chat-scene} & 55.5 & 50.2 & 57.1 & 52.4 \\

\midrule
\multicolumn{5}{l}{\textcolor{gray}{\textit{Vision LLM Based Model}}} \vspace{1mm} \\

\rowcolor{green!5} Qwen2-VL-7B~\cite{Qwen2VL} & 5.4 & 5.1 & 21.1 & 19.9 \\

\rowcolor{green!7} Qwen2-VL-7B (GPT4Scene)  & 40.5$\,_\text{\textcolor{darkgray}{+35.1}}$ & 36.7$\,_\text{\textcolor{darkgray}{+31.6}}$ & 45.4$\,_\text{\textcolor{darkgray}{+24.3}}$ & 42.1$\,_\text{\textcolor{darkgray}{+22.2}}$ \\

\rowcolor{green!10} Qwen2-VL-7B (GPT4Scene)-HD  & 50.9$\,_\text{\textcolor{darkgray}{+45.5}}$ & 46.4$\,_\text{\textcolor{darkgray}{+41.3}}$ & 53.7$\,_\text{\textcolor{darkgray}{+32.6}}$ & 50.0$\,_\text{\textcolor{darkgray}{+30.1}}$ \\

\rowcolor{green!15} Qwen2-VL-7B (GPT4Scene)-HDM  & \textbf{62.6}$\,_\text{\textcolor{darkgray}{+51.9}}$ & \textbf{57.0}$\,_\text{\textcolor{darkgray}{+32.3}}$ & \textbf{64.5}$\,_\text{\textcolor{darkgray}{+43.4}}$ & \textbf{59.8}$\,_\text{\textcolor{darkgray}{+39.9}}$ \\

\bottomrule
\end{tabular}

        }
        \vspace{-1mm}
        \caption{\textbf{Evaluation of 3D Visual Grounding on ScanRefer~\cite{Scanrefer} and Multi3DRef~\cite{Multi3DRefer}. }
        Our method reaches SOTA performance over all methods.}
        \label{tab_visual_grounding}
    \end{minipage}
    \\
    \hspace{-4mm}
    \begin{minipage}{0.47\textwidth}
    \centering
        \vspace{1mm}
        \tablestyle{7pt}{1.17}
        \resizebox{\linewidth}{!}{
        \begin{tabular}{lcccc}
\toprule
3D QA GPT Score & \multicolumn{3}{c}{Combat Situation} & \multicolumn{1}{c}{GPT Score} \\
\cmidrule(lr){2-4} \cmidrule(lr){5-5}
Methods & Win & Tie & Lose & Score \\
\midrule

\multicolumn{5}{l}{\textcolor{gray}{\textit{VLM vs 3D LLM}}} \vspace{1mm} \\

\rowcolor{green!5} Qwen2-VL-7B~\cite{Qwen2VL} {\small vs} Chat-Scene~\cite{Chat-scene} & 74 & 243 & 683 & 465 \\

\rowcolor{green!7} Qwen2-VL-7B(GPT4Scene) {\small vs} Chat-Scene & 543 & 145 & 312 & 1774 \\

\bottomrule
\end{tabular}

        }
        \vspace{-1mm}
        \caption{\textbf{GPT Score evaluation on 3D question answering.} We evaluated the GPT Score on ScanQA~\cite{scanqa}, comparing our model’s outputs with Chat-Scene~\cite{Chat-scene}. The fine-tuned model achieved significantly higher GPT Scores than current SOTA methods.
        }
        \label{tab_gpt_score}
    \end{minipage}
    \vspace{-4mm}
\end{table}

\vspace{2mm}
\noindent \textbf{Qualitative Results. }
We present qualitative results in \Cref{fig_qualitative_results}, conducted in a zero-shot setting on GPT-4o. Stitched frames provide an overview of the scene, while individual frames capture details and actions. Beyond standard tasks like object captioning, spatial description, and counting, GPT4Scene handles embodied tasks, such as directing users to retrieve paper from a nearby bookshelf. In the last row, pink-highlighted segments indicate BEV images as input, enhancing navigation capabilities. GPT4Scene also excels in navigation and patrolling tasks, performing industrial inspections by observing machine indicators.

\vspace{2mm}
\noindent \textbf{GPT  Score: New Evaluation Way. }
We propose a novel GPT Score for 3D QA assessment, comparing two methods with the state-of-the-art 3D LLM Chat-Scene~\cite{Chat-scene} as baseline. In the comparison with Qwen2-VL-7B~\cite{Qwen2VL}, we input both models' outputs and the ground truth into GPT-4o~\cite{gpt4o} to evaluate which response is better. We select 1,000 questions from ScanQA and define "win," "tie," and "lose" to indicate whether Qwen2-VL-7B's answer is superior to, equivalent to, or inferior to Chat-Scene's~\cite{Chat-scene}. These outcomes are assigned scores of 3, 1, and -1 points respectively (analogous to football scoring), which are aggregated to derive a comprehensive GPT Score. Our results show that Qwen2-VL-7B (without fine-tuning) underperforms Chat-Scene, while its fine-tuned version (GPT4Scene) significantly outperforms it, demonstrating its effectiveness.

\begin{table*}[t]
    \vspace{-5mm}
    \centering
    \label{tab_overall}
    \begin{minipage}{0.32\textwidth}
        \centering
        \tablestyle{12pt}{1.2}
        \resizebox{\linewidth}{!}{
            \begin{tabular}{lc}
\toprule
\multirow{2}{*}{\textbf{Small Objects Sets} } & ScanQA \\
\cmidrule(lr){2-2}
& ROUGE \\
\midrule

Chat-Scene & 37.5  \\
GPT-4o (GPT4Scene)  & 35.4   \\
Qwen2-VL-7B (GPT4Scene)  & 39.4   \\

\bottomrule
\end{tabular}
        }
        \vspace{1mm}
        \subcaption{\textbf{\footnotesize Ablation on Small Objects.}}
        \label{tab_small_objects_sets}
    \end{minipage}
    \hfill
    \begin{minipage}{0.32\textwidth}
        \centering
        \tablestyle{12pt}{1.2}
        \resizebox{\linewidth}{!}{
            \begin{tabular}{lc}
\toprule
\multirow{2}{*}{\makecell{\textbf{Precision and Size} \\ \textbf{of STO-markers}}}& ScanQA \\
\cmidrule(lr){2-2}
& ROUGE \\
\midrule

Qwen2-VL-7B (GPT4Scene)  & 43.6  \\
 - Delete 30\% STO-markers  & 42.7   \\
 - Adaptive size adjustment  & 43.0  \\

\bottomrule
\end{tabular}
        }
        \vspace{1mm}
        \subcaption{\textbf{\footnotesize Ablation on STO-markers.}}
        \label{tab_sto_markers}
    \end{minipage}
    \hfill
    \begin{minipage}{0.32\textwidth}
        \centering
        \tablestyle{11pt}{1.15}
        \resizebox{\linewidth}{!}{
            \begin{tabular}{lc}
\toprule
\multirow{2}{*}{\textbf{BEV Reconstruction Quality} } & ScanQA \\
\cmidrule(lr){2-2}
& ROUGE \\
\midrule

Qwen2-VL-7B (GPT4Scene) & 43.6  \\
 - SLAM3R~\cite{SLAM3R}, 50-frame intervals  & 42.4  \\
 - SLAM3R~\cite{SLAM3R}, 100-frame intervals & 41.9 \\
 - SLAM3R~\cite{SLAM3R}, 200-frame intervals & 43.2 \\

\bottomrule
\end{tabular}
        }
        \vspace{1mm}
        \subcaption{\textbf{\footnotesize Ablation on BEV Reconstruction Quality.}}
        \label{tab_bev_reconstruction}
    \end{minipage}
    \vspace{-1mm}
    \caption{\textbf{Overall Performance Comparison.} We investigated GPT4Scene's performance on small objects and its robustness in handling STO-markers and BEV reconstruction quality, demonstrating that our model exhibits strong performance and robustness.} 
    \label{tab_overall_performance_comparison}
\end{table*}

\begin{table}[t]
    \vspace{-3mm}
    \begin{minipage}{0.47\textwidth}
    \centering
        \tablestyle{8.5pt}{1.2}
        \resizebox{\linewidth}{!}{
        \begin{tabular}{lcccc}
\toprule
\multirow{2}{*}{Ablation} & \multicolumn{2}{c}{ScanQA} & \multicolumn{2}{c}{Multi3DRef@0.5} \\
\cmidrule(lr){2-3} \cmidrule(lr){4-5}
& ROUGE & CIDEr & MT & ALL \\
\midrule
Qwen2-VL-7B (GPT4Scene) & 43.6 & 90.9 & 36.3  & 42.1 \\
w/o BEV Image  & 42.3 & 87.1 & 27.8  & 32.1 \\
w/o STO markers  & 42.8 & 88.4 & -  & - \\
w/o BEV Image \& STO markers & 41.7 & 85.0 & - & - \\

\bottomrule
\end{tabular}

        }
        \vspace{-2mm}
        \caption{\textbf{Ablation Study on our modules.} Removing BEV images or STO-markers leads to a performance decrease, and further removal of STO-markers causes an additional decrease.}
        \label{tab_abalation}
    \end{minipage}
    \vspace{-3mm}
\end{table}

\newpage
\subsection{Ablation Study}
\label{sec_ablation_study}
~~~~In this section, we conduct ablation studies to validate the effectiveness of GPT4Scene. First, we evaluate its robustness, including performance on small objects, followed by analyzing the robustness of STO-markers and reconstruction quality. Next, we perform module-wise ablation to assess individual components. Finally, we investigate the impact of frame intervals and resolution settings.

\vspace{2mm}
\noindent \textbf{Ablation on Robustness Analysis.}
We tested the performance of GPT4Scene on small objects by selecting 1,000 instances, such as cups and towels. The results demonstrate that the fine-tuned Qwen2-VL-7B~\cite{Qwen2VL} model with GPT4Scene exhibits significantly improved performance. Following this, we analyzed the robustness of STO-markers. Two modifications were implemented: first, removing 30\% of the markers, and second, dynamically adjusting marker sizes based on object dimensions. Experimental results indicate that both operations had minimal impact on overall performance, confirming the robustness of STO-markers. The third experiment investigated the impact of BEV (Bird’s Eye View) map reconstruction quality on results. Originally, we used BundleFusion~\cite{BundleFusion} from ScanNet~\cite{ScanNet} for reconstruction. We replaced this with SLAM3R~\cite{SLAM3R}, a real-time reconstruction method, and tested settings with frame intervals of 50, 100, and 200. The results show that while reconstruction quality slightly affects performance, the influence is negligible. This suggests that the reconstructed BEV primarily serves to provide global scene context rather than precise geometric details.

\vspace{2mm}
\noindent \textbf{Ablation on Our Modules.}
\Cref{tab_zero_shot_setting} demonstrates that BEV images and STO-markers enhance spatial understanding under zero-shot settings. We further validate this observation in the fine-tuning scenario through evaluations on both 3D question answering and 3D visual grounding tasks. It should be noted that STO-markers are essential for visual grounding——when they are removed, validation can only be performed on the question answering task. As shown in \Cref{tab_abalation}, both BEV images and STO-markers contribute significantly to the performance improvements. 

\vspace{2mm}
\noindent \textbf{Ablation on Frames and Resolution.}
Here, we conduct ablation studies on additional factors using 3D question answering and visual grounding as benchmark tasks. The results are shown in~\Cref{tab_abalation_2}. We define three configurations: the base mode with 8 frames and longer side resolution of 128, the HD (high-resolution) mode with 8 frames at 512 resolution, and the HDM (high-resolution multi-frame) mode with 32 frames at 512 resolution. The first three rows demonstrate that image resolution substantially impacts visual grounding performance while providing limited improvement for the QA task. The final three rows reveal that frame quantity augmentation enhances indoor scene comprehension, showing more pronounced gains in grounding performance compared to marginal QA improvements.

\begin{table}[t]
    \vspace{-3mm}
    \begin{minipage}{0.47\textwidth}
    \centering
        \tablestyle{12pt}{1}
        \resizebox{\linewidth}{!}{
        \begin{tabular}{clcccc}
\toprule
{Num} & \multirow{2}{*}{Resolution} & \multicolumn{2}{c}{ScanQA} & \multicolumn{2}{c}{ScanRefer}  \\
\cmidrule(lr){3-4} \cmidrule(lr){5-6}
Frames & & ROUGE & CIDEr & Acc\@0.25 & Acc\@0.5 \\
\midrule
\multirow{3}{*}{8} & 128 (base) & 43.6 & 90.9 & 40.5 & 36.7 \\
& 256 & 43.8 & 90.0 & 49.2 & 44.8 \\
& 512 (HD) & 43.6 & 89.9 & 50.9 & 46.4 \\
\midrule
16 & 512 & 45.4 & 93.4 & 58.6 & 53.4 \\
\midrule
\rowcolor{gray!10}32 & 512 (HDM) & 46.5 & 96.3 & 62.6 &57.0 \\
\bottomrule
\end{tabular}

        }
        \vspace{-2mm}
        \caption{\textbf{Ablation Study on frames and resolution.} Results show that the number of frames affects both QA and grounding, while resolution significantly impacts grounding.}
        \label{tab_abalation_2}
        \vspace{-3mm}
    \end{minipage}
    \vspace{-2mm}
\end{table}

\section{Conclusion}
\label{sec_conclusion}
~~~~We propose GPT4Scene, a framework enhancing Vision-Language Models (VLMs) to understand 3D scenes directly from visual inputs. By reconstructing 3D point clouds for Bird’s Eye View (BEV) images and aligning video frames with spatial-temporal object (STO) markers, we enable global-local scene comprehension. GPT4Scene achieves state-of-the-art 3D QA performance with zero-shot GPT-4o and fine-tuned smaller VLMs (e.g., Qwen2-VL) using our ScanAlign dataset. Fine-tuned models even excel with raw video inputs, proving effective 3D understanding.

\appendix
\section{Prompts of Closed-source VLMs}
\label{sec_closed-sourse_prompts}

\begin{figure*}[t]
  \vspace{-1mm}
  \centering
  \includegraphics[width=0.97\textwidth]{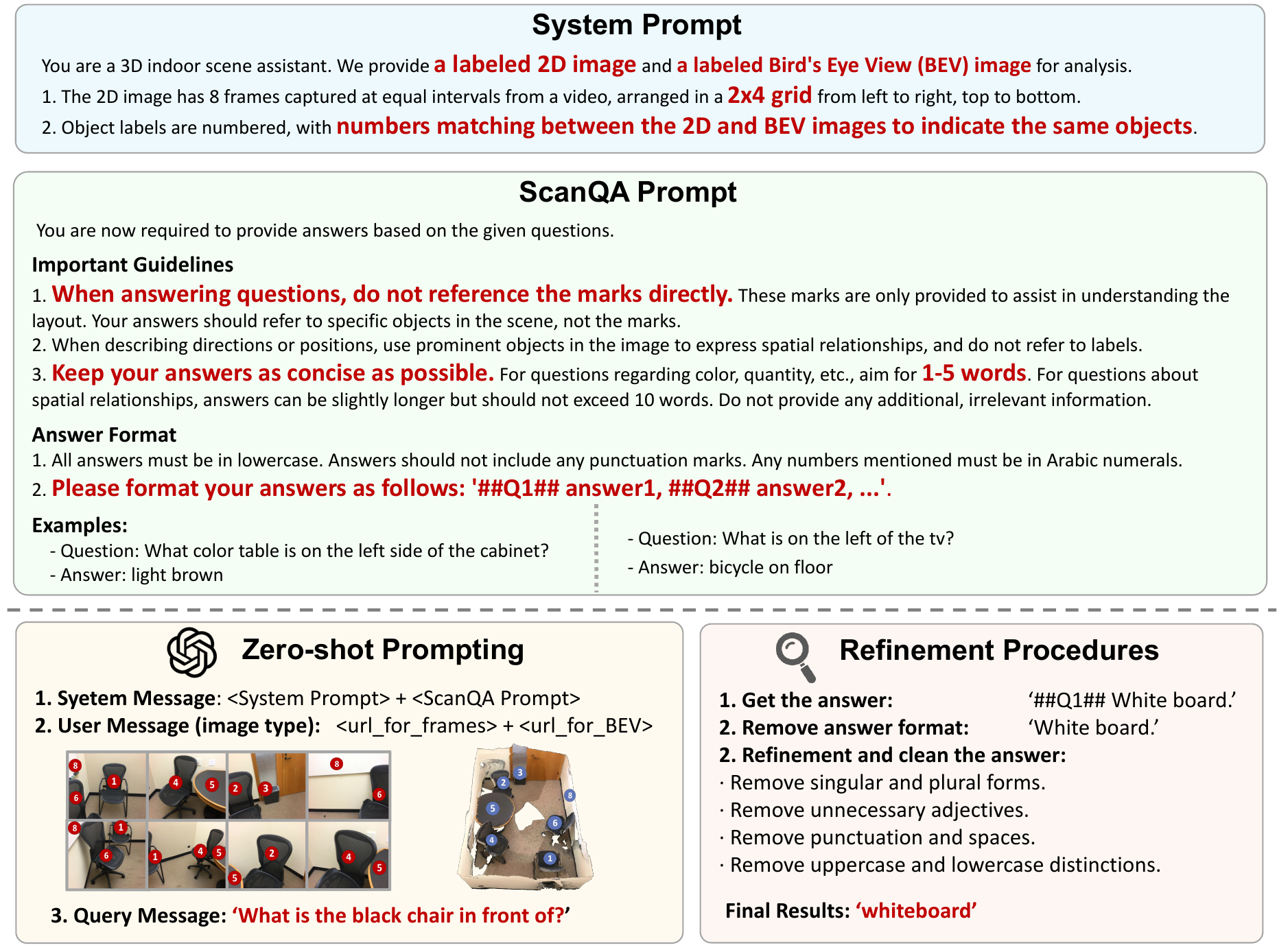}
  \caption{\textbf{Prompts of Closed-sourse VLMs. } We show the prompts used for GPT-4o (GPT4Scene), which consist of a system prompt and a benchmark prompt. After generating responses, we further refine them.}
  \label{fig_closed_source_prompts}
  \vspace{-2mm}
\end{figure*}

Here, we present the prompts used by GPT4Scene in the closed-source VLMs (GPT-4o), as illustrated in \Cref{fig_closed_source_prompts}. The process begins with a general \textbf{system prompt}, which outlines the overview of two images provided as input. The first image is a stitched 2D view captured from a video, with dimensions of $2 \times 4$. The second image represents a \textbf{BEV} (Bird's Eye View). Subsequently, we perform evaluations across various tasks and benchmarks, with each benchmark associated with a specific prompt. We take the 3D question-answering task on \textbf{ScanQA} as an example. The benchmark prompt consists of three parts:

\begin{enumerate}[label=\arabic*.]
    \item \textbf{Important Guidelines:} It clarifies that while we provide object IDs for reference, they cannot be directly used when answering questions. Additionally, it specifies adapting the response style to match that of ScanQA. Since ScanQA's responses are typically short single words, we aim to keep the answers concise within the benchmark prompt, targeting 1-5 words.
    
    \item \textbf{Answer Format:} In this part, we use a standardized regularized format to structure the answers, which helps improve accuracy when addressing questions.
    
    \item \textbf{Examples:} we include two example cases.
\end{enumerate}

Our \textbf{zero-shot prompting process} is illustrated in the bottom-left corner \Cref{fig_closed_source_prompts}. The system prompt and ScanQA prompt are used as the system message. In the user message, we input the stitched image and BEV image. Finally, the query message includes the question. The responses generated through this process require further refinement, as depicted in the bottom-right corner of \Cref{fig_closed_source_prompts}. First, we remove the regularized formatting from the answers. Next, we clean the answers by addressing singular/plural forms and case sensitivity. This final step ensures that we obtain the refined answers.

\begin{table}[t]
    \hspace{-3mm}
    \begin{minipage}{0.47\textwidth}
    \centering
        \tablestyle{15pt}{1.02}
        \resizebox{\linewidth}{!}{
        \begin{tabular}{lcc}
\toprule
\multirow{2}{*}{Task type} & \multicolumn{2}{c}{Model} \\
\cmidrule(lr){2-3}
& Qwen2-VL & Ours \\
\midrule
Action Sequence               & 85.5                & 82.0               \\ 
Action Prediction             & 69.5                & \textbf{70.5}               \\ 
Action Antonym                & 83.0                & \textbf{86.0}               \\ 
Fine-grained Action           & 51.5                & 51.5               \\ 
Unexpected Action             & 82.0                & 78.0               \\ 
Object Existence              & 87.5                & \textbf{88.5}               \\ 
Object Interaction            & 82.0                & 81.5               \\ 
Object Shuffle                & 41.0                & \textbf{45.0}               \\ 
Moving Direction              & 42.0                & 40.0               \\ 
Action Localization           & 65.0                & \textbf{66.5}               \\ 
Scene Transition              & 93.5                & \textbf{94.0}               \\ 
Action Count                  & 47.5                & 43.5               \\ 
Moving Count                  & 69.5                & 71.5               \\ 
Moving Attribute              & 90.0                & 88.5               \\ 
State Change                  & 48.0                & \textbf{49.0}               \\ 
Fine-grained Pose             & 63.0                & 63.5               \\ 
Character Order               & 74.5                & 71.0               \\ 
Egocentric Navigation         & 39.5                & \textbf{41.5}               \\ 
Episodic Reasoning            & 47.0                & 47.0               \\ 
Counterfactual Inference      & 62.5                & \textbf{65.5}               \\ 
\midrule
\textbf{Avg}                  & \textbf{66.2}       & \textbf{66.225}    \\ 
\bottomrule
\end{tabular}


        }
        \vspace{-2mm}
        \caption{\textbf{The result of MVBench \cite{MVBench}.} After fine-tuning with ScanAlign in GPT4Scene, our model shows improved 2D understanding, particularly in object and action metrics.}
        \label{tab_2d_benchmark}
        \vspace{-3mm}
    \end{minipage}
\end{table}

\begin{table}[t]
    \hspace{-3mm}
    \begin{minipage}{0.46\textwidth}
    \centering
        \tablestyle{19pt}{0.98}
        \resizebox{\linewidth}{!}{
        
\begin{tabular}{lcc}
\toprule
\multirow{2}{*}{Benchmark} & \multicolumn{2}{c}{Model} \\
\cmidrule(lr){2-3}
& Qwen2-VL & Ours \\
\midrule
MMBench-EN$_{val}$ \cite{MMBench}          & 82.4           & 81.2          \\ 
MMBench-CN$_{val}$ \cite{MMBench}         & 81.7           & 79.9          \\ 
MMStar \cite{MMStar}                    & 60.7           & 57.6          \\ 
RealWorldQA \cite{RealWorldQA}               & 70.1           & 68.5          \\ 
\midrule
Video-MME \cite{videomme}                  & 59.8           & 58.4          \\ 

\bottomrule
\end{tabular}
        }
        \vspace{-2mm}
        \caption{\textbf{The result of 2D Multi-modal Benchmark.} After fine-tuning with ScanAlign in GPT4Scene, our model's 2D understanding capabilities did not decline.}
        \label{tab_2d_all}
        \vspace{-4mm}
    \end{minipage}
\end{table}

\begin{table*}[!t]
    \begin{subtable}{0.95\linewidth}
    \centering
    \tablestyle{15pt}{1.2}
    \resizebox{\linewidth}{!}{
        \begin{tabular}{lcccccccc}
\toprule
\multirow{2}{*}{Methods} & & \multicolumn{4}{c}{\textbf{BLEU-n Metrics}} & \multicolumn{3}{c}{\textbf{Language Generation Metrics}} \\
\cmidrule(lr){3-6}\cmidrule(lr){7-9}
 & EM-1 & BLEU-1 & BLEU-2 & BLEU-3 & BLEU-4 & ROUGE & METEOR & CIDEr \\
\midrule
\multicolumn{4}{l}{\textcolor{gray}{\textit{Task-Specific Model}}} \vspace{1mm} \\
ScanQA~\cite{scanqa} & 21.1 & 30.2 & 20.4 & 15.1 & 10.1 & 33.3 & 13.1 & 64.9 \\

3D-VLP~\cite{3dvlp} & 21.7 & 30.5 & 21.3 & 16.7 & 11.2 & 34.5 & 13.5 & 67.0 \\

3D-Vista~\cite{3D-VisTA} & -- & -- & -- & -- & 13.9 & 35.7 & -- & -- \\

\midrule
\multicolumn{4}{l}{\textcolor{gray}{\textit{3D LLM Based Model}}} \vspace{1mm} \\

3D-LLM~\cite{3dllm} & 20.5 & 39.3 & 25.2 & 18.4 & 12.0 & 35.7 & 14.5 & 69.4 \\

LL3DA~\cite{ll3da} & -- & -- & -- & -- & 13.5 & 37.3 & 15.9 & 76.8 \\

LEO~\cite{LEO} & 24.5 & -- & -- & -- & 11.5 & 39.3 & 16.2 & 80.0 \\

Scene-LLM~\cite{scenellm} & 27.2 & 43.6 & 26.8 & 19.1 & 12.0 & 40.0 & 16.6 & 80.0 \\

Chat-Scene~\cite{Chat-scene} & 21.6 & 43.2 & 29.1 & 20.6 & 14.3 & 41.6 & 18.0 & 87.7 \\

\midrule
\multicolumn{4}{l}{\textcolor{gray}{\textit{Vision LLM Based Model}}} \vspace{1mm} \\



\rowcolor{green!5} Qwen2-VL-7B~\cite{Qwen2VL} & 19.0 & 27.8 & 13.6 & 6.3 & 3.0 & 29.3 & 11.4 & 53.9 \\

\rowcolor{green!7} Qwen2-VL-7B (GPT4Scene) & 25.5 & 43.4 & 29.3 & 20.9 & 14.6 & 43.6 & 17.7 & 90.9 \\

\rowcolor{green!10} Qwen2-VL-7B (GPT4Scene)-HD & 26.3 & 41.9 & 28.6 & 21.6 & \textbf{15.9} & 43.6 & 17.6 & 89.9 \\

\rowcolor{green!15} Qwen2-VL-7B (GPT4Scene)-HDM & \textbf{28.2} & \textbf{44.4} & \textbf{30.3} & \textbf{22.3} & 15.5 & \textbf{46.5} & \textbf{18.9} & \textbf{96.3} \\

\bottomrule
\end{tabular}
    }
    \end{subtable}
    \caption{
        \textbf{Full Evaluation of 3D Question Answering on ScanQA~\cite{scanqa}.}
    }
    \label{tab_3d_scanqa}
    \vspace{3mm}  
\end{table*}

\begin{table*}[!t]
    \begin{subtable}{0.95\linewidth}
    \centering
    \tablestyle{18pt}{1.2}
    \resizebox{\linewidth}{!}{
        \begin{tabular}{lcccccccc}
\toprule
\multirow{2}{*}{Methods} & \multicolumn{6}{c}{\textbf{Test Set}} &
 \multirow{2}{*}{Avg.(EM-1)} & \multirow{2}{*}{EM-R1} \\
\cmidrule(lr){2-7}
 & What & Is & How & Can & Which & Others &
\\
\midrule
\multicolumn{5}{l}{\textcolor{gray}{\textit{Task-Specific Model}}} \vspace{1mm} \\
ClipBERT~\cite{sqa3d}  & 30.2 & 60.1 & 38.7 & 63.3 & 42.5 & 42.7 & 43.3 & --  \\

SQA3D~\cite{sqa3d} & 31.6 & 63.8 & 46.0 & 69.5 & 43.9 & 45.3 & 46.6 & --  \\

3D-VisTA~\cite{3D-VisTA} & 34.8 & 63.3 & 45.4 & 69.8 & 47.2 & 48.1 & 48.5 & -- \\

\midrule
\multicolumn{5}{l}{\textcolor{gray}{\textit{3D LLM Based Model}}} \vspace{1mm} \\

PQ3D~\cite{PQ3D} & 37.1 & 61.3 & 44.5 & 60.9 & 47.0 & 45.1 & 47.1 & -- \\

LEO~\cite{LEO} & -- & -- & -- & -- & -- & -- & 50.0 & 52.4 \\

Scene-LLM~\cite{scenellm} &  40.9 & 69.1 & 45.0 & \textbf{70.8} & 47.2 & 52.3 & 54.2 & -- \\

Chat-Scene~\cite{Chat-scene} & 45.4 & 67.0 & 52.0 & 69.5 & 49.9 & 55.0 & 54.6 & 57.5 \\

\midrule
\multicolumn{5}{l}{\textcolor{gray}{\textit{Vision LLM Based Model}}} \vspace{1mm} \\



\rowcolor{green!5} Qwen2-VL-7B~\cite{Qwen2VL} & 29.0 & 59.2 & 33.4 & 50.5 & 44.2 & 43.2 & 40.7 & 46.7 \\

\rowcolor{green!7} Qwen2-VL-7B (GPT4Scene) & 50.7 & \textbf{70.9} & 48.0 & 70.5 & 52.9 & 59.3 & 57.4 & 60.7 \\

\rowcolor{green!10} Qwen2-VL-7B (GPT4Scene)-HD & 51.4 & 69.1 & 50.2 & 69.4 & 51.3 & 57.9 & 57.2 & 60.4 \\

\rowcolor{green!15} Qwen2-VL-7B (GPT4Scene)-HDM & \textbf{55.9} & 69.9 & \textbf{50.8} & 68.7 & \textbf{53.3} & \textbf{60.4} & \textbf{59.4} & \textbf{62.4} \\

\bottomrule
\end{tabular}
    }
    \end{subtable}
    \caption{
        \textbf{Full Evaluation of 3D Question Answering on SQA3D~\cite{sqa3d}.}
    }
    \label{tab_3d_sqa3d}
    \vspace{3mm}  
\end{table*}

\vspace{-1mm}
\section{2D Multi-modal Benchmark}
We tested the fine-tuned \textit{Qwen2-VL-7B (GPT4Scene)} model on 2D image and video multimodal large models. \Cref{tab_2d_benchmark} shows the results of MVBench. As we can see, our model shows improvement for the object and action metrics, indicating that the model fine-tuned with ScanAlign is better at handling spatial variations and the information of objects in the scene. \Cref{tab_2d_all} presents the results on other benchmarks, where we can observe that after our fine-tuning, the model's ability to understand images and videos did not decline significantly. It demonstrates the effectiveness of ScanAlign.

\vspace{-1mm}
\section{Qualitative Results}
\begin{figure*}[t]
  \vspace{-1mm}
  \centering
  \includegraphics[width=\textwidth]{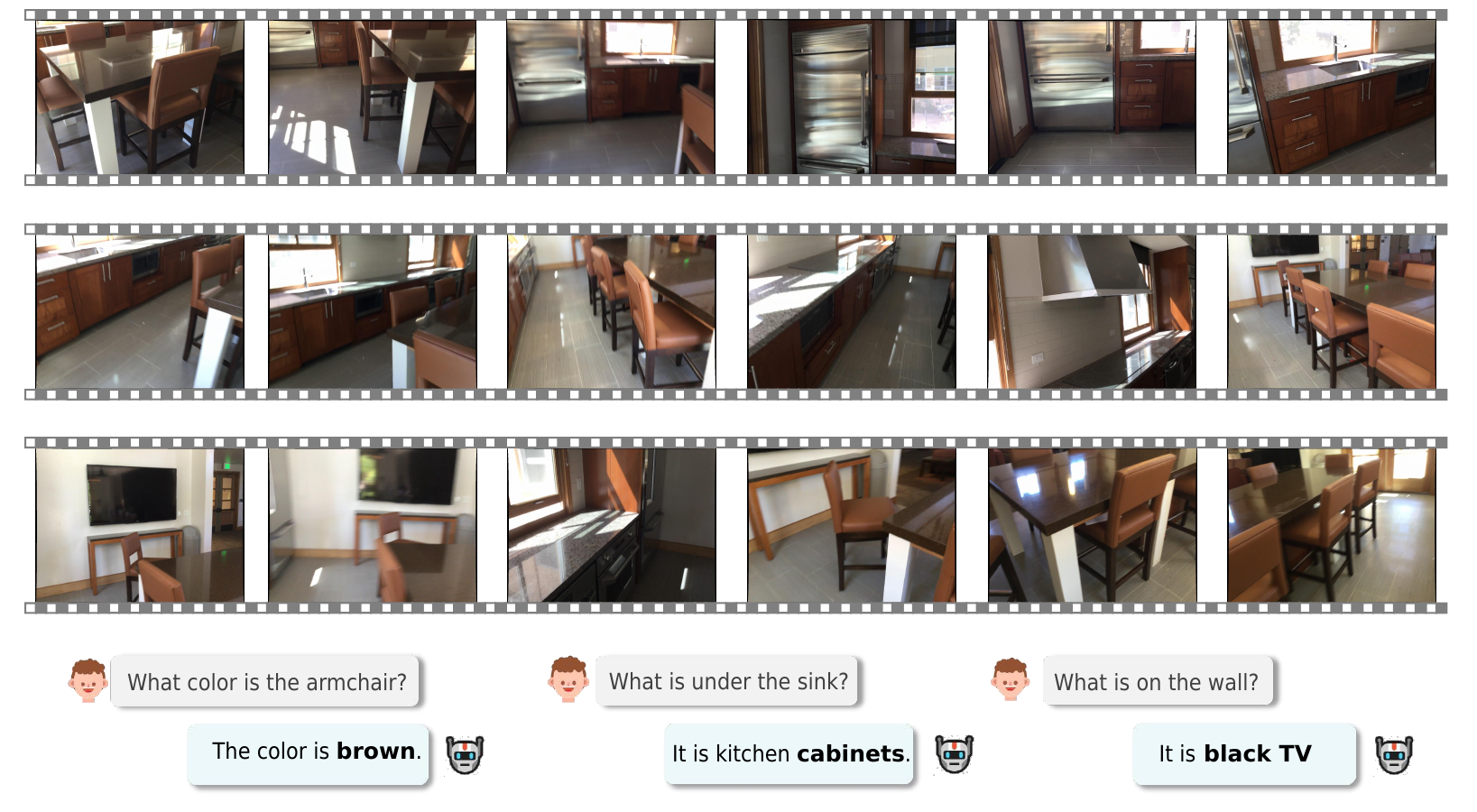}
  \caption{\textbf{Qualitive Results: Question Answering. }We provide videos without object annotations.}
  \label{fig_sup1}
  \vspace{3mm}
\end{figure*}

\begin{figure*}[t]
  \vspace{-1mm}
  \centering
  \includegraphics[width=\textwidth]{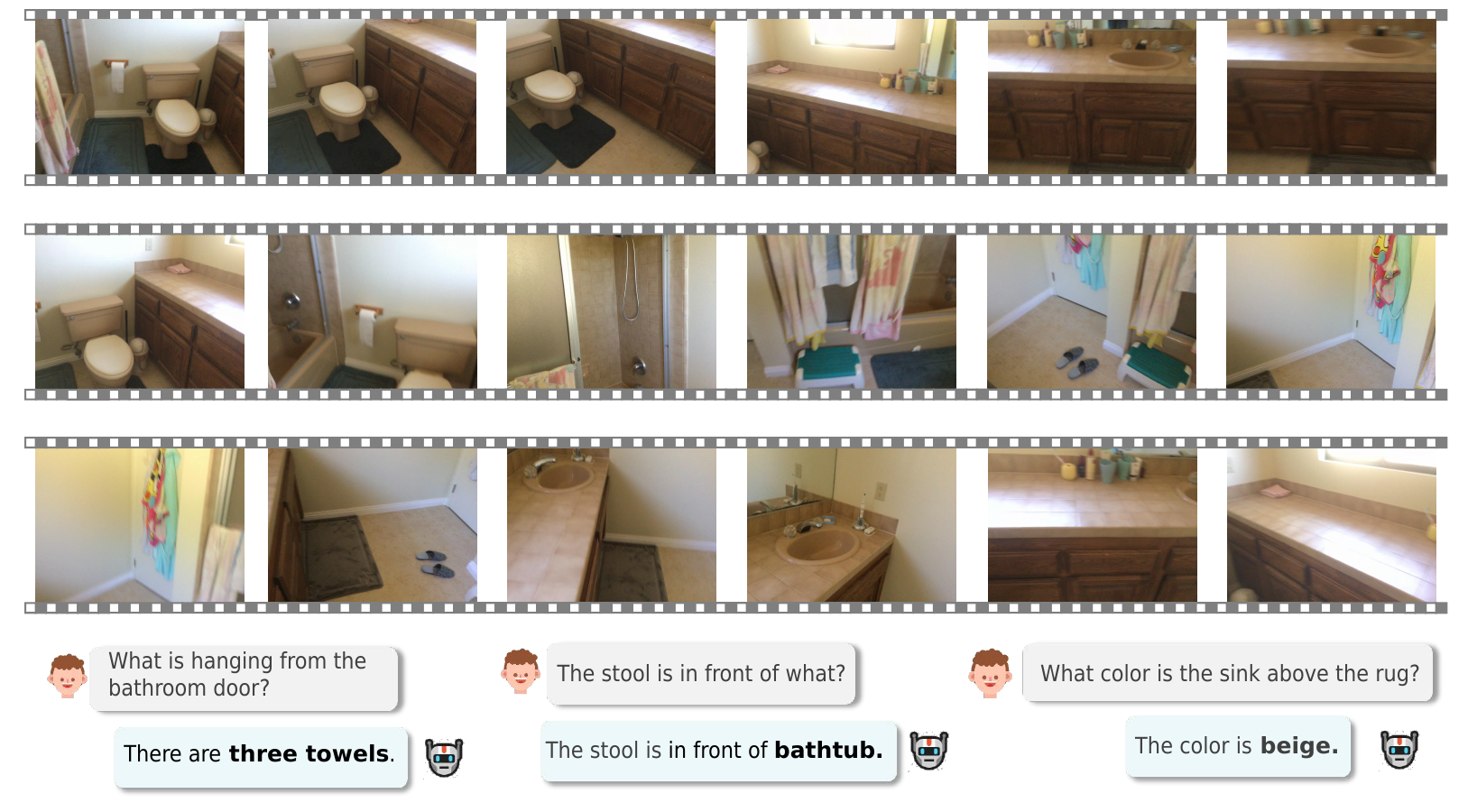}
  \caption{\textbf{Qualitive Results: Question Answering. }We provide videos without object annotations.}
  \label{fig_sup2}
  \vspace{4mm}
\end{figure*}

\begin{figure*}[t]
  \vspace{-1mm}
  \centering
  \includegraphics[width=\textwidth]{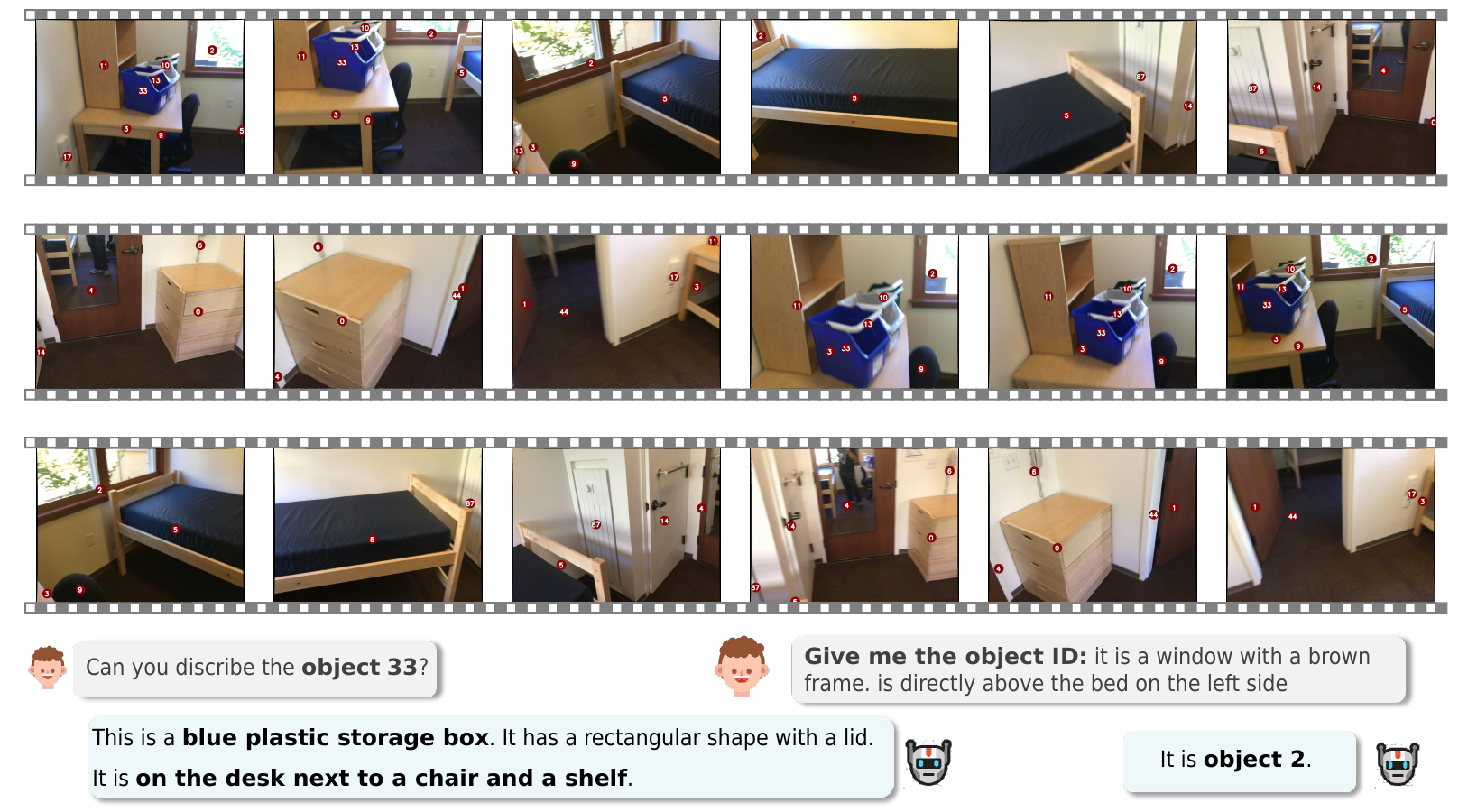}
  \caption{\textbf{Qualitive Results: 3D Dense Caption and Visual Grounding. }We provide videos with object annotations.}
  \label{fig_sup3}
  \vspace{3mm}
\end{figure*}

\begin{figure*}[t]
  \vspace{-1mm}
  \centering
  \includegraphics[width=\textwidth]{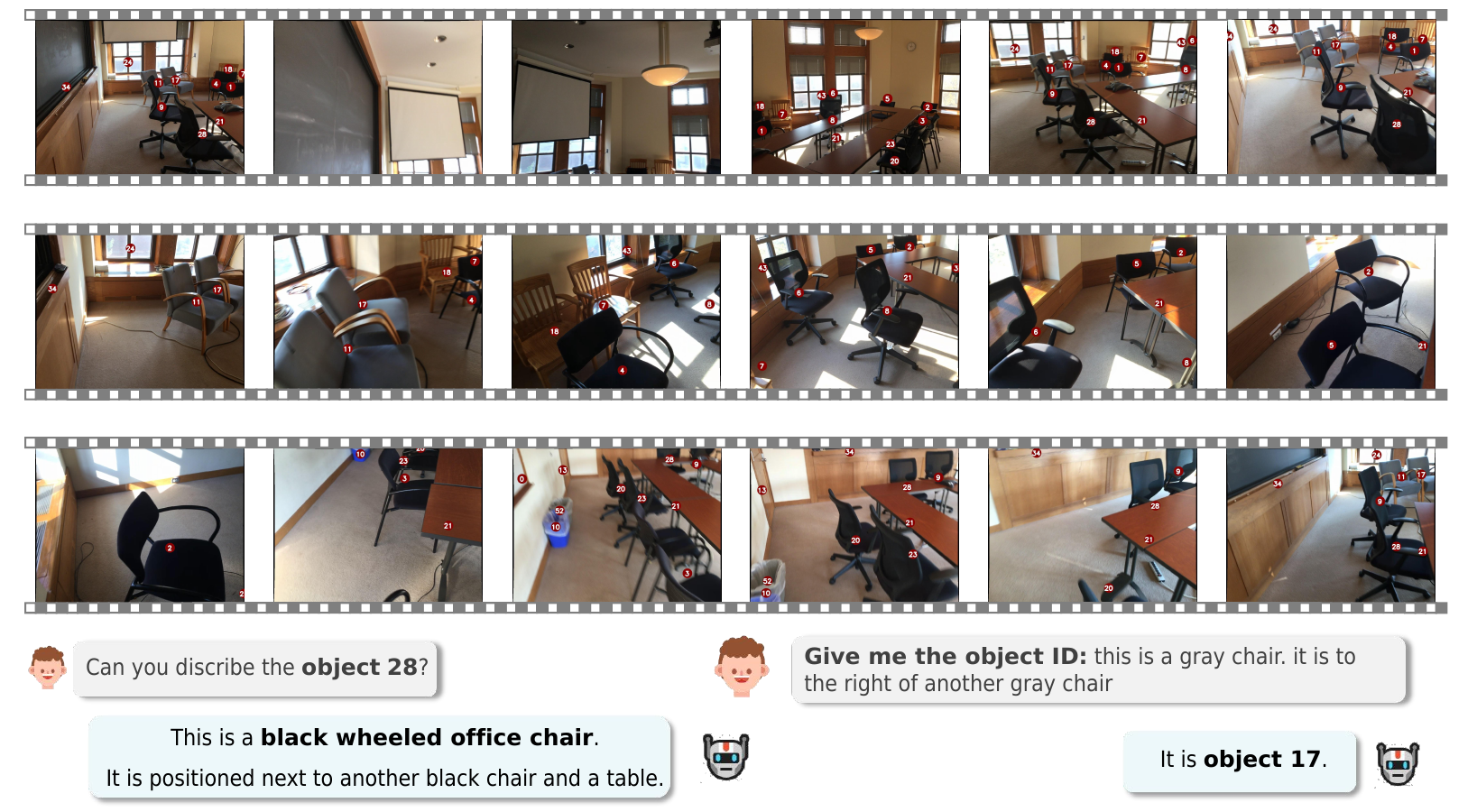}
  \caption{\textbf{Qualitive Results: 3D Dense Caption and Visual Grounding. }We provide videos with object annotations.}
  \label{fig_sup4}
  \vspace{4mm}
\end{figure*}

\Cref{fig_sup1,fig_sup2,fig_sup3,fig_sup4} present our qualitative results obtained from Qwen2-VL after fine-tuning on ScanAlign. \Cref{fig_sup1,fig_sup2} demonstrate 3D question answering performance using only unannotated video inputs as context. Our observations indicate the model's capability to effectively address queries from both ScanQA~\cite{scanqa} and SQA3D~\cite{sqa3d} benchmarks. Meanwhile, \Cref{fig_sup3,fig_sup4} exhibit the model's competence in 3D dense captioning and 3D visual grounding tasks requiring annotated inputs. These outcomes substantiate that our framework can generate precise responses through visual information processing alone, without dependency on 3D point cloud inputs. Here we emphasize that post-training with the GPT4Scene framework, the model achieves accurate QA performance using only pure video input, thereby demonstrating GPT4Scene's effectiveness in enhancing visual comprehension capabilities.

\vspace{-1mm}
\section{Full Quantitive Results}
Here, we present the complete metrics for all five benchmarks. \Cref{tab_3d_scanqa} and \Cref{tab_3d_sqa3d} show results for ScanQA~\cite{scanqa} and SQA3D~\cite{sqa3d}. \Cref{tab_3d_scan2cap} provides the full results for Scan2Cap~\cite{Scan2cap}, while \Cref{tab_3d_scanrefer} and \Cref{tab_3d_multiref3d} present the results for ScanRefer~\cite{Scanrefer} and Multi3DRef~\cite{Multi3DRefer}. Our model significantly improves across all benchmarks, highlighting that only pure vision language models can understand 3D scenes effectively.



\begin{table*}[!t]
    \begin{subtable}{0.95\linewidth}
    \centering
    \vspace{-3mm}
    \tablestyle{13pt}{1.3}
    \resizebox{\linewidth}{!}{
        \begin{tabular}{lcccc|cccc}
\toprule
\multirow{2}{*}{Methods} & \multicolumn{4}{c}{\textbf{IoU@0.25}} & \multicolumn{4}{c}{\textbf{IoU@0.5}}
\\\cmidrule(lr){2-5}\cmidrule(lr){6-9}
& CIDEr & BLEU-4 & METEOR & ROUGE & CIDEr & BLEU-4 & METEOR & ROUGE
\\
\midrule
\multicolumn{5}{l}{\textcolor{gray}{\textit{Task-Specific Model}}} \vspace{1mm} \\
Scan2Cap~\cite{Scan2cap} & 56.8 & 34.2 & 26.3 & 55.3 & 39.1 & 23.3 & 22.0 & 44.5 \\

3DJCG~\cite{3djcg} & 64.7 & 40.2 & 27.7 & 59.2 & 49.5 & 31.0 & 24.2 & 50.8 \\

X-Trans2Cap~\cite{X-trans2cap} & 61.8 & 35.7 & 26.6 & 54.7 & 43.9 & 25.1 & 22.5 & 45.3 \\

D3Net~\cite{D3Net} & -- & -- & -- & -- & 62.6 & 35.7 & 25.7 & 53.9 \\

3D-VLP~\cite{3dvlp} & 70.7 & 41.0 & 28.1 & 59.7 & 54.9 & 32.3 & 24.8 & 51.5 \\

Vote2Cap-DETR~\cite{vote2cap-detr} & 71.5 & 39.3 & 28.3 & 59.3 & 62.6 & 35.7 & 25.7 & 53.9 \\

3D-VisTA~\cite{3D-VisTA} & 71.0 & 36.5 & 28.4 & 57.6 & 66.9 & 34.0 & 27.1 & 54.3 \\

\midrule
\multicolumn{5}{l}{\textcolor{gray}{\textit{3D LLM Based Model}}} \vspace{1mm} \\

LL3DA~\cite{ll3da} & 74.2 & 41.4 & 27.8 & 59.5 & 65.2 & 36.8 & 26.0 & 55.1 \\

LEO~\cite{LEO} & -- & -- & -- & -- & 68.4 & 36.9 & 27.7 & 57.8 \\

Chat-Scene~\cite{Chat-scene} & 81.9 & 38.2 & 29.0 & 60.6 & 77.2 & 36.3 & 28.0 & 58.1 \\

Robin3D~\cite{Robin3D} & -- & -- & -- & -- & \textbf{87.2} & 38.4 & -- & -- \\

\midrule
\multicolumn{5}{l}{\textcolor{gray}{\textit{Vision LLM Based Model}}} \vspace{1mm} \\

\rowcolor{green!5} Qwen2-VL-7B~\cite{Qwen2VL} & 0.0 & 3.8 & 16.7 & 24.7 & 0.0 &3.8 & 16.5 & 24.6 \\

\rowcolor{green!7} Qwen2-VL-7B (GPT4Scene) & 63.8 & 36.3 & 26.5 & 57.6 & 60.6 & 34.2 & 25.6 & 55.2 \\

\rowcolor{green!10} Qwen2-VL-7B (GPT4Scene)-HD & 79.1 & 40.4 & 28.3 & 60.2 & 74.4 & 37.9 & 27.3 & 57.7 \\

\rowcolor{green!15} Qwen2-VL-7B (GPT4Scene)-HDM & \textbf{91.7} & \textbf{43.1} & \textbf{29.3} & \textbf{61.9} & 86.3 & \textbf{40.6} & \textbf{28.2} & \textbf{59.3} \\

\bottomrule
\end{tabular}

    }
    \end{subtable}
    \caption{
        \textbf{Full Evaluation of 3D Dense Caption on Scan2Cap~\cite{Scan2cap}.}
    }
    \label{tab_3d_scan2cap}
    \vspace{-1mm}  
\end{table*}

\begin{table*}[!t]
    \begin{subtable}{0.95\linewidth}
    \centering
    \vspace{-3mm}
    \tablestyle{15pt}{1.2}
    \resizebox{\linewidth}{!}{
        \begin{tabular}{lcccc|cc}
\toprule
\multirow{2}{*}{Methods} & \multicolumn{2}{c}{\textbf{Unique}} & \multicolumn{2}{c}{\textbf{Multiple}} & \multicolumn{2}{c}{\textbf{Overall}}
\\ \cmidrule(lr){2-3} \cmidrule(lr){4-5} \cmidrule(lr){6-7}
 & Acc@0.25 & Acc@0.5 & Acc@0.25 & Acc@0.5 & Acc@0.25 & Acc@0.5 
\\
\midrule
\multicolumn{5}{l}{\textcolor{gray}{\textit{Task-Specific Model}}} \vspace{1mm} \\

ScanRefer~\cite{Scanrefer} & 76.3 & 53.5 & 32.7 & 21.1 & 41.2 & 27.4 \\

TGNN~\cite{tgnn} & 68.6 & 56.8 & 29.8 & 23.2 & 37.4 & 29.7 \\

SAT~\cite{X-trans2cap} & 73.2 & 50.8 & 37.6 & 25.2 & 44.5 & 30.1  \\

InstanceRefer~\cite{Instancerefer} & 75.7 & 64.7 & 29.4 & 23.0 & 38.4 & 31.1 \\

3DVG-Transformer~\cite{3DVG-Transformer} & 81.9 & 60.6 & 39.3 & 28.4 & 47.6 & 34.7 \\

MVT~\cite{MVT} & 77.7 & 66.4 & 31.9 & 25.3 & 40.8 & 33.3 \\

3D-SPS~\cite{3D-SPS} & 84.1 & 66.7 & 40.3 & 29.8 & 48.8 & 37.0 \\

ViL3DRel~\cite{ViL3DRel} & 81.6 & 68.6 & 40.3 & 30.7 & 47.9 & 37.7 \\

3DJCG~\cite{3djcg} & 83.5 & 64.3 & 41.4 & 30.8 & 49.6 & 37.3 \\

D3Net~\cite{D3Net} & -- & 72.0 & -- & 30.1 & -- & 37.9 \\

BUTD-DETR~\cite{BUTD} & 84.2 & 66.3 & 46.6 & 35.1 & 52.2 & 39.8 \\

HAM~\cite{HAM} & 79.2 & 67.9 & 41.5 & 34.0 & 48.8 & 40.6 \\

3DRP-Net~\cite{3DRP-Net} & 83.1 & 67.7 & 42.1 & 32.0 & 50.1 & 38.9 \\

3D-VLP~\cite{3dvlp} & 84.2 & 64.6 & 43.5 & 33.4 & 51.4 & 39.5 \\

EDA~\cite{EDA} & 85.8 & 68.6 & \textbf{49.1} & 37.6 & 54.6 & 42.3 \\

M3DRef-CLIP~\cite{Multi3DRefer} & 85.3 & 77.2 & 43.8 & 36.8 & 51.9 & 44.7 \\

3D-VisTA~\cite{3D-VisTA} & 81.6 & 75.1 & 43.7 & 39.1 & 50.6 & 45.8 \\

ConcreteNet~\cite{ConcreteNet} & 86.4 & 82.1 & 42.4 & 38.4 & 50.6 & 46.5 \\

DOrA~\cite{DORA} & -- & -- & -- & -- & 52.8 & 44.8  \\

\midrule
\multicolumn{5}{l}{\textcolor{gray}{\textit{3D LLM Based Model}}} \\

Chat-Scene~\cite{Chat-scene} & 89.6 & 82.5 & 47.8 & 42.9 & 55.5 & 50.2 \\

Robin3D~\cite{Robin3D} & -- & -- & -- & -- & 60.8 & 55.1 \\

\midrule
\multicolumn{5}{l}{\textcolor{gray}{\textit{Vision LLM Based Model}}} \\

\rowcolor{green!5} Qwen2-VL-7B~\cite{Qwen2VL} & 6.3 & 6.3 & 5.1 & 4.8 & 5.4 & 5.1 \\

\rowcolor{green!7} Qwen2-VL-7B (GPT4Scene) & 65.5 & 61.2 & 34.8 & 31.1 & 40.5 & 36.7 \\

\rowcolor{green!10} Qwen2-VL-7B (GPT4Scene)-HD & 77.5 & 71.9 & 44.9 & 40.6 & 50.9 & 46.4 \\

\rowcolor{green!15} Qwen2-VL-7B (GPT4Scene)-HDM & \textbf{90.3} & \textbf{83.7} & \textbf{56.4} & \textbf{50.9} & \textbf{62.6} & \textbf{57.0} \\

\bottomrule
\end{tabular}

    }
    \end{subtable}
    \caption{
        \textbf{Full Evaluation of 3D Visual Grounding on ScanRefer~\cite{Scanrefer}.}
    }
    \label{tab_3d_scanrefer}
    \vspace{-1mm}  
\end{table*}

\begin{table*}[!t]
    \begin{subtable}{0.95\linewidth}
    \centering
    \vspace{-3mm}
    \tablestyle{7pt}{1.3}
    \resizebox{\linewidth}{!}{
        \begin{tabular}{lcccccccccc}
\toprule
\multirow{2}{*}{Methods} &  ZT w/o D & ZT w/ D & \multicolumn{2}{c}{ST w/o D} & \multicolumn{2}{c}{ST w/ D} & \multicolumn{2}{c}{MT} & \multicolumn{2}{c}{ALL}
\\
\cmidrule(lr){2-3}\cmidrule(lr){4-5}\cmidrule(lr){6-7}\cmidrule(lr){8-9}\cmidrule(lr){10-11}
 & F1 & F1 & F1@0.25 & F1@0.5 & F1@0.25 & F1@0.5 & F1@0.25 & F1@0.5 & F1@0.25 & F1@0.5
\\

\midrule
\multicolumn{5}{l}{\textcolor{gray}{\textit{Task-Specific Model}}} \vspace{1mm} \\

3DVG-Trans+~\cite{3DVG-Transformer} & 87.1 & 45.8 & -- & 27.5 & -- & 16.7 & -- & 26.5 & -- & 25.5 \\

D3Net (Grounding)~\cite{D3Net} & 81.6 & 32.5 & -- & 38.6 & -- & 23.3 & -- & 35.0 & -- & 32.2 \\

3DJCG (Grounding)~\cite{3djcg} & 94.1 & 66.9 & -- & 26.0 & -- & 16.7 & -- & 26.2 & -- & 26.6 \\

M3DRef-CLIP~\cite{Multi3DRefer} & 81.8 & 39.4 & 53.5 & 47.8 & 34.6 & 30.6 & 43.6 & 37.9 & 42.8 & 38.4 \\

\midrule
\multicolumn{5}{l}{\textcolor{gray}{\textit{3D LLM Based Model}}} \vspace{1mm} \\

Chat-Scene~\cite{Chat-scene} & 90.3 & 62.6 & 82.9  & 75.9 & 49.1 & 44.5 & 45.7 & 41.1 & 57.1 & 52.4 \\

\midrule
\multicolumn{5}{l}{\textcolor{gray}{\textit{Vision LLM Based Model}}} \vspace{1mm} \\

\rowcolor{green!5} Qwen2-VL-7B~\cite{Qwen2VL} & 84.8 & 58.5 & 20.1 & 19.1 & 14.7 & 13.5 & 16.8 & 15.4 & 21.1 & 19.9 \\

\rowcolor{green!7} Qwen2-VL-7B (GPT4Scene) & 85.2 & 61.4 & 60.1 & 55.1 & 37.7 & 34.4 & 39.4 & 36.3 & 45.4 & 42.1 \\

\rowcolor{green!10} Qwen2-VL-7B (GPT4Scene)-HD & 93.6 & 81.8 & 72.5 & 66.2 & 46.6 & 42.9 & 41.8 & 38.9 & 53.7 & 50.0 \\

\rowcolor{green!15} Qwen2-VL-7B (GPT4Scene)-HDM & \textbf{97.4} & \textbf{84.4} & \textbf{85.0} & \textbf{77.7} & \textbf{59.9} & \textbf{55.1} & \textbf{48.6} & \textbf{44.6} & \textbf{64.5} & \textbf{59.8} \\

\bottomrule
\end{tabular}

    }
    \end{subtable}
    \caption{
        \textbf{Full Evaluation of 3D Visual Grounding on Multi3DRef~\cite{Multi3DRefer}.}
    }
    \label{tab_3d_multiref3d}
    \vspace{-1mm}  
\end{table*}

\clearpage

\clearpage

\clearpage

{
    \small
    \begin{thebibliography}{128}
\providecommand{\natexlab}[1]{#1}
\providecommand{\url}[1]{\texttt{#1}}
\expandafter\ifx\csname urlstyle\endcsname\relax
  \providecommand{\doi}[1]{doi: #1}\else
  \providecommand{\doi}{doi: \begingroup \urlstyle{rm}\Url}\fi

\bibitem[Abdelreheem et~al.(2024)Abdelreheem, Olszewski, Lee, Wonka, and Achlioptas]{Scanents3d}
Ahmed Abdelreheem, Kyle Olszewski, Hsin-Ying Lee, Peter Wonka, and Panos Achlioptas.
\newblock Scanents3d: Exploiting phrase-to-3d-object correspondences for improved visio-linguistic models in 3d scenes.
\newblock In \emph{WACV}, 2024.

\bibitem[Achlioptas et~al.(2020)Achlioptas, Abdelreheem, Xia, Elhoseiny, and Guibas]{Referit3d}
Panos Achlioptas, Ahmed Abdelreheem, Fei Xia, Mohamed Elhoseiny, and Leonidas Guibas.
\newblock Referit3d: Neural listeners for fine-grained 3d object identification in real-world scenes.
\newblock In \emph{ECCV}, 2020.

\bibitem[Alayrac et~al.(2022)Alayrac, Donahue, Luc, Miech, Barr, Hasson, Lenc, Mensch, Millican, Reynolds, et~al.]{Flamingo}
Jean-Baptiste Alayrac, Jeff Donahue, Pauline Luc, Antoine Miech, Iain Barr, Yana Hasson, Karel Lenc, Arthur Mensch, Katherine Millican, Malcolm Reynolds, et~al.
\newblock Flamingo: a visual language model for few-shot learning.
\newblock In \emph{NeurIPS}, 2022.

\bibitem[Ataallah et~al.(2024)Ataallah, Shen, Abdelrahman, Sleiman, Zhu, Ding, and Elhoseiny]{Minigpt4-video}
Kirolos Ataallah, Xiaoqian Shen, Eslam Abdelrahman, Essam Sleiman, Deyao Zhu, Jian Ding, and Mohamed Elhoseiny.
\newblock Minigpt4-video: Advancing multimodal llms for video understanding with interleaved visual-textual tokens.
\newblock \emph{arXiv:2404.03413}, 2024.

\bibitem[Azuma et~al.(2022)Azuma, Miyanishi, Kurita, and Kawanabe]{scanqa}
Daichi Azuma, Taiki Miyanishi, Shuhei Kurita, and Motoaki Kawanabe.
\newblock Scanqa: 3d question answering for spatial scene understanding.
\newblock In \emph{CVPR}, 2022.

\bibitem[Bakr et~al.(2022)Bakr, Alsaedy, and Elhoseiny]{LAR}
Eslam Bakr, Yasmeen Alsaedy, and Mohamed Elhoseiny.
\newblock Look around and refer: 2d synthetic semantics knowledge distillation for 3d visual grounding.
\newblock In \emph{NeurIPS}, 2022.

\bibitem[Baruch et~al.(2021)Baruch, Chen, Dehghan, Dimry, Feigin, Fu, Gebauer, Joffe, Kurz, Schwartz, et~al.]{Arkitscenes}
Gilad Baruch, Zhuoyuan Chen, Afshin Dehghan, Tal Dimry, Yuri Feigin, Peter Fu, Thomas Gebauer, Brandon Joffe, Daniel Kurz, Arik Schwartz, et~al.
\newblock Arkitscenes: A diverse real-world dataset for 3d indoor scene understanding using mobile rgb-d data.
\newblock In \emph{NeurIPS}, 2021.

\bibitem[Cai et~al.(2022)Cai, Zhao, Zhang, Sheng, and Xu]{3djcg}
Daigang Cai, Lichen Zhao, Jing Zhang, Lu Sheng, and Dong Xu.
\newblock 3djcg: A unified framework for joint dense captioning and visual grounding on 3d point clouds.
\newblock In \emph{CVPR}, 2022.

\bibitem[Chandrasegaran et~al.(2024)Chandrasegaran, Gupta, Hadzic, Kota, He, Eyzaguirre, Durante, Li, Wu, and Fei-Fei]{HourVideo}
Keshigeyan Chandrasegaran, Agrim Gupta, Lea~M. Hadzic, Taran Kota, Jimming He, Cristóbal Eyzaguirre, Zane Durante, Manling Li, Jiajun Wu, and Li Fei-Fei.
\newblock Hourvideo: 1-hour video-language understanding.
\newblock \emph{arXiv:2411.04998}, 2024.

\bibitem[Chang et~al.(2017)Chang, Dai, Funkhouser, Halber, Niessner, Savva, Song, Zeng, and Zhang]{Matterport3d}
Angel Chang, Angela Dai, Thomas Funkhouser, Maciej Halber, Matthias Niessner, Manolis Savva, Shuran Song, Andy Zeng, and Yinda Zhang.
\newblock Matterport3d: Learning from rgb-d data in indoor environments.
\newblock In \emph{3DV}, 2017.

\bibitem[Changpinyo et~al.(2021)Changpinyo, Sharma, Ding, and Soricut]{CC12M}
Soravit Changpinyo, Piyush Sharma, Nan Ding, and Radu Soricut.
\newblock Conceptual 12m: Pushing web-scale image-text pre-training to recognize long-tail visual concepts.
\newblock In \emph{CVPR}, 2021.

\bibitem[Chen et~al.(2020)Chen, Chang, and Nie{\ss}ner]{Scanrefer}
Dave~Zhenyu Chen, Angel~X Chang, and Matthias Nie{\ss}ner.
\newblock Scanrefer: 3d object localization in rgb-d scans using natural language.
\newblock In \emph{ECCV}, 2020.

\bibitem[Chen et~al.(2022{\natexlab{a}})Chen, Wu, Nie{\ss}ner, and Chang]{D3Net}
Dave~Zhenyu Chen, Qirui Wu, Matthias Nie{\ss}ner, and Angel~X Chang.
\newblock D3net: a speaker-listener architecture for semi-supervised dense captioning and visual grounding in rgb-d scans.
\newblock In \emph{ECCV}, 2022{\natexlab{a}}.

\bibitem[Chen et~al.(2022{\natexlab{b}})Chen, Luo, Wei, Ma, and Zhang]{HAM}
Jing Chen, Wenhao Luo, Xizhou Wei, Lin Ma, and Wei Zhang.
\newblock Ham: Hierarchical attention model with high performance for 3d visual grounding.
\newblock \emph{arXiv:2210.12513}, 2022{\natexlab{b}}.

\bibitem[Chen et~al.(2023{\natexlab{a}})Chen, Zhu, Shen, Li, Liu, Zhang, Krishnamoorthi, Chandra, Xiong, and Elhoseiny]{Minigpt-v2}
Jun Chen, Deyao Zhu, Xiaoqian Shen, Xiang Li, Zechun Liu, Pengchuan Zhang, Raghuraman Krishnamoorthi, Vikas Chandra, Yunyang Xiong, and Mohamed Elhoseiny.
\newblock Minigpt-v2: large language model as a unified interface for vision-language multi-task learning.
\newblock \emph{arXiv:2310.09478}, 2023{\natexlab{a}}.

\bibitem[Chen et~al.(2024{\natexlab{a}})Chen, Li, Dong, Zhang, Zang, Chen, Duan, Wang, Qiao, Lin, et~al.]{MMStar}
Lin Chen, Jinsong Li, Xiaoyi Dong, Pan Zhang, Yuhang Zang, Zehui Chen, Haodong Duan, Jiaqi Wang, Yu Qiao, Dahua Lin, et~al.
\newblock Are we on the right way for evaluating large vision-language models?
\newblock In \emph{NeurIPS}, 2024{\natexlab{a}}.

\bibitem[Chen et~al.(2022{\natexlab{c}})Chen, Guhur, Tapaswi, Schmid, and Laptev]{3DMV-VQA}
Shizhe Chen, Pierre-Louis Guhur, Makarand Tapaswi, Cordelia Schmid, and Ivan Laptev.
\newblock Language conditioned spatial relation reasoning for 3d object grounding.
\newblock In \emph{NeurIPS}, 2022{\natexlab{c}}.

\bibitem[Chen et~al.(2022{\natexlab{d}})Chen, Guhur, Tapaswi, Schmid, and Laptev]{ViL3DRel}
Shizhe Chen, Pierre-Louis Guhur, Makarand Tapaswi, Cordelia Schmid, and Ivan Laptev.
\newblock Language conditioned spatial relation reasoning for 3d object grounding.
\newblock In \emph{NeurIPS}, 2022{\natexlab{d}}.

\bibitem[Chen et~al.(2023{\natexlab{b}})Chen, Zhu, Chen, Lei, Yu, and Chen]{vote2cap-detr}
Sijin Chen, Hongyuan Zhu, Xin Chen, Yinjie Lei, Gang Yu, and Tao Chen.
\newblock End-to-end 3d dense captioning with vote2cap-detr.
\newblock In \emph{CVPR}, 2023{\natexlab{b}}.

\bibitem[Chen et~al.(2024{\natexlab{b}})Chen, Chen, Zhang, Li, Yu, Fei, Zhu, Fan, and Chen]{ll3da}
Sijin Chen, Xin Chen, Chi Zhang, Mingsheng Li, Gang Yu, Hao Fei, Hongyuan Zhu, Jiayuan Fan, and Tao Chen.
\newblock Ll3da: Visual interactive instruction tuning for omni-3d understanding reasoning and planning.
\newblock In \emph{CVPR}, 2024{\natexlab{b}}.

\bibitem[Chen et~al.(2021)Chen, Gholami, Nie{\ss}ner, and Chang]{Scan2cap}
Zhenyu Chen, Ali Gholami, Matthias Nie{\ss}ner, and Angel~X Chang.
\newblock Scan2cap: Context-aware dense captioning in rgb-d scans.
\newblock In \emph{CVPR}, 2021.

\bibitem[Chen et~al.(2023{\natexlab{c}})Chen, Wu, Wang, Su, Chen, Xing, Zhong, Zhang, Zhu, Lu, Li, Luo, Lu, Qiao, and Dai]{InternVL}
Zhe Chen, Jiannan Wu, Wenhai Wang, Weijie Su, Guo Chen, Sen Xing, Muyan Zhong, Qinglong Zhang, Xizhou Zhu, Lewei Lu, Bin Li, Ping Luo, Tong Lu, Yu Qiao, and Jifeng Dai.
\newblock Internvl: Scaling up vision foundation models and aligning for generic visual-linguistic tasks.
\newblock \emph{arXiv:2312.14238}, 2023{\natexlab{c}}.

\bibitem[Cheng et~al.(2024)Cheng, Leng, Zhang, Xin, Li, Chen, Zhu, Zhang, Luo, Zhao, et~al.]{Video-llama2}
Zesen Cheng, Sicong Leng, Hang Zhang, Yifei Xin, Xin Li, Guanzheng Chen, Yongxin Zhu, Wenqi Zhang, Ziyang Luo, Deli Zhao, et~al.
\newblock Videollama 2: Advancing spatial-temporal modeling and audio understanding in video-llms.
\newblock \emph{arXiv:2406.07476}, 2024.

\bibitem[Dai et~al.(2017{\natexlab{a}})Dai, Chang, Savva, Halber, Funkhouser, and Nie{\ss}ner]{ScanNet}
Angela Dai, Angel~X. Chang, Manolis Savva, Maciej Halber, Thomas Funkhouser, and Matthias Nie{\ss}ner.
\newblock Scannet: Richly-annotated 3d reconstructions of indoor scenes.
\newblock In \emph{CVPR}, 2017{\natexlab{a}}.

\bibitem[Dai et~al.(2017{\natexlab{b}})Dai, Nie{\ss}ner, Zoll{\"o}fer, Izadi, and Theobalt]{BundleFusion}
Angela Dai, Matthias Nie{\ss}ner, Michael Zoll{\"o}fer, Shahram Izadi, and Christian Theobalt.
\newblock Bundlefusion: Real-time globally consistent 3d reconstruction using on-the-fly surface re-integration.
\newblock \emph{TOG}, 2017{\natexlab{b}}.

\bibitem[Dai et~al.(2023)Dai, Li, Li, Tiong, Zhao, Wang, Li, Fung, and Hoi]{InstructBLIP}
Wenliang Dai, Junnan Li, Dongxu Li, Anthony Meng~Huat Tiong, Junqi Zhao, Weisheng Wang, Boyang Li, Pascale Fung, and Steven Hoi.
\newblock Instructblip: Towards general-purpose vision-language models with instruction tuning.
\newblock \emph{arXiv:2305.06500}, 2023.

\bibitem[Deitke et~al.(2022)Deitke, VanderBilt, Herrasti, Weihs, Ehsani, Salvador, Han, Kolve, Kembhavi, and Mottaghi]{ProcTHOR}
Matt Deitke, Eli VanderBilt, Alvaro Herrasti, Luca Weihs, Kiana Ehsani, Jordi Salvador, Winson Han, Eric Kolve, Aniruddha Kembhavi, and Roozbeh Mottaghi.
\newblock Procthor: Large-scale embodied ai using procedural generation.
\newblock In \emph{NeurIPS}, 2022.

\bibitem[Ding et~al.(2023)Ding, Yang, Xue, Zhang, Bai, and Qi]{PLA}
Runyu Ding, Jihan Yang, Chuhui Xue, Wenqing Zhang, Song Bai, and Xiaojuan Qi.
\newblock Pla: Language-driven open-vocabulary 3d scene understanding.
\newblock In \emph{CVPR}, 2023.

\bibitem[Dong et~al.(2024)Dong, Zhang, Zang, Cao, Wang, Ouyang, Wei, Zhang, Duan, Cao, Zhang, Li, Yan, Gao, Zhang, Li, Li, Chen, He, Zhang, Qiao, Lin, and Wang]{internlmxcomposer2}
Xiaoyi Dong, Pan Zhang, Yuhang Zang, Yuhang Cao, Bin Wang, Linke Ouyang, Xilin Wei, Songyang Zhang, Haodong Duan, Maosong Cao, Wenwei Zhang, Yining Li, Hang Yan, Yang Gao, Xinyue Zhang, Wei Li, Jingwen Li, Kai Chen, Conghui He, Xingcheng Zhang, Yu Qiao, Dahua Lin, and Jiaqi Wang.
\newblock Internlm-xcomposer2: Mastering free-form text-image composition and comprehension in vision-language large model.
\newblock \emph{arXiv:2401.16420}, 2024.

\bibitem[Fu et~al.(2024{\natexlab{a}})Fu, Dai, Luo, Li, Ren, Zhang, Wang, Zhou, Shen, Zhang, et~al.]{videomme}
Chaoyou Fu, Yuhan Dai, Yondong Luo, Lei Li, Shuhuai Ren, Renrui Zhang, Zihan Wang, Chenyu Zhou, Yunhang Shen, Mengdan Zhang, et~al.
\newblock Video-mme: The first-ever comprehensive evaluation benchmark of multi-modal llms in video analysis.
\newblock \emph{arXiv:2405.21075}, 2024{\natexlab{a}}.

\bibitem[Fu et~al.(2024{\natexlab{b}})Fu, Liu, Chen, Nie, and Xiong]{scenellm}
Rao Fu, Jingyu Liu, Xilun Chen, Yixin Nie, and Wenhan Xiong.
\newblock Scene-llm: Extending language model for 3d visual understanding and reasoning.
\newblock \emph{arXiv:2403.11401}, 2024{\natexlab{b}}.

\bibitem[Guo et~al.(2023{\natexlab{a}})Guo, Tang, Zhang, Wang, Wang, Zhao, and Li]{viewrefer}
Zoey Guo, Yiwen Tang, Ray Zhang, Dong Wang, Zhigang Wang, Bin Zhao, and Xuelong Li.
\newblock Viewrefer: Grasp the multi-view knowledge for 3d visual grounding.
\newblock In \emph{ICCV}, 2023{\natexlab{a}}.

\bibitem[Guo et~al.(2023{\natexlab{b}})Guo, Zhang, Zhu, Tang, Ma, Han, Chen, Gao, Li, Li, and Heng]{Point-Bind}
Ziyu Guo, Renrui Zhang, Xiangyang Zhu, Yiwen Tang, Xianzheng Ma, Jiaming Han, Kexin Chen, Peng Gao, Xianzhi Li, Hongsheng Li, and Pheng-Ann Heng.
\newblock Point-bind \& point-llm: Aligning point cloud with multi-modality for 3d understanding, generation, and instruction following.
\newblock \emph{arXiv:2309.00615}, 2023{\natexlab{b}}.

\bibitem[Ha and Song(2022)]{Semantic-abstraction}
Huy Ha and Shuran Song.
\newblock Semantic abstraction: Open-world 3d scene understanding from 2d vision-language models.
\newblock In \emph{CORL}, 2022.

\bibitem[He et~al.(2021)He, Zhao, Luo, Hui, Huang, Zhang, and Liu]{Transrefer3d}
Dailan He, Yusheng Zhao, Junyu Luo, Tianrui Hui, Shaofei Huang, Aixi Zhang, and Si Liu.
\newblock Transrefer3d: Entity-and-relation aware transformer for fine-grained 3d visual grounding.
\newblock In \emph{ACMMM}, 2021.

\bibitem[Hegde et~al.(2023)Hegde, Valanarasu, and Patel]{Clip-goes-3d}
Deepti Hegde, Jeya Maria~Jose Valanarasu, and Vishal Patel.
\newblock Clip goes 3d: Leveraging prompt tuning for language grounded 3d recognition.
\newblock In \emph{ICCV}, 2023.

\bibitem[Hong et~al.(2023)Hong, Zhen, Chen, Zheng, Du, Chen, and Gan]{3dllm}
Yining Hong, Haoyu Zhen, Peihao Chen, Shuhong Zheng, Yilun Du, Zhenfang Chen, and Chuang Gan.
\newblock 3d-llm: Injecting the 3d world into large language models.
\newblock In \emph{NeurIPS}, 2023.

\bibitem[Huang et~al.(2023{\natexlab{a}})Huang, Wang, Huang, Liu, Cheng, Zhao, Jin, and Zhao]{Chat-3Dv2}
Haifeng Huang, Zehan Wang, Rongjie Huang, Luping Liu, Xize Cheng, Yang Zhao, Tao Jin, and Zhou Zhao.
\newblock Chat-3d v2: Bridging 3d scene and large language models with object identifiers.
\newblock \emph{arXiv:2312.08168}, 2023{\natexlab{a}}.

\bibitem[Huang et~al.(2024{\natexlab{a}})Huang, Chen, Wang, Huang, Xu, Wang, Liu, Cheng, Zhao, Pang, et~al.]{Chat-scene}
Haifeng Huang, Yilun Chen, Zehan Wang, Rongjie Huang, Runsen Xu, Tai Wang, Luping Liu, Xize Cheng, Yang Zhao, Jiangmiao Pang, et~al.
\newblock Chat-scene: Bridging 3d scene and large language models with object identifiers.
\newblock In \emph{NeurIPS}, 2024{\natexlab{a}}.

\bibitem[Huang et~al.(2023{\natexlab{b}})Huang, Yong, Ma, Linghu, Li, Wang, Li, Zhu, Jia, and Huang]{LEO}
Jiangyong Huang, Silong Yong, Xiaojian Ma, Xiongkun Linghu, Puhao Li, Yan Wang, Qing Li, Song-Chun Zhu, Baoxiong Jia, and Siyuan Huang.
\newblock An embodied generalist agent in 3d world.
\newblock \emph{arXiv:2311.12871}, 2023{\natexlab{b}}.

\bibitem[Huang et~al.(2021)Huang, Lee, Chen, and Liu]{tgnn}
Pao-Hsiang Huang, Hsin-Hsi Lee, Hwann-Tzong Chen, and Tyng-Luh Liu.
\newblock Text-guided graph neural networks for referring 3d instance segmentation.
\newblock In \emph{AAAI}, 2021.

\bibitem[Huang et~al.(2022{\natexlab{a}})Huang, Chen, Jia, and Wang]{MVT}
Shijia Huang, Yilun Chen, Jiaya Jia, and Liwei Wang.
\newblock Multi-view transformer for 3d visual grounding.
\newblock In \emph{CVPR}, 2022{\natexlab{a}}.

\bibitem[Huang et~al.(2022{\natexlab{b}})Huang, Chen, Jia, and Wang]{MVT-3DVG}
Shijia Huang, Yilun Chen, Jiaya Jia, and Liwei Wang.
\newblock Multi-view transformer for 3d visual grounding.
\newblock In \emph{CVPR}, 2022{\natexlab{b}}.

\bibitem[Huang et~al.(2024{\natexlab{b}})Huang, Dong, Wang, Hao, Singhal, Ma, Lv, Cui, Mohammed, Patra, et~al.]{kosmos}
Shaohan Huang, Li Dong, Wenhui Wang, Yaru Hao, Saksham Singhal, Shuming Ma, Tengchao Lv, Lei Cui, Owais~Khan Mohammed, Barun Patra, et~al.
\newblock Language is not all you need: Aligning perception with language models.
\newblock In \emph{NeurIPS}, 2024{\natexlab{b}}.

\bibitem[Jain et~al.(2022)Jain, Gkanatsios, Mediratta, and Fragkiadaki]{BUTD}
Ayush Jain, Nikolaos Gkanatsios, Ishita Mediratta, and Katerina Fragkiadaki.
\newblock Bottom up top down detection transformers for language grounding in images and point clouds.
\newblock In \emph{ECCV}, 2022.

\bibitem[Jia et~al.(2024)Jia, Chen, Yu, Wang, Niu, Liu, Li, and Huang]{SceneVerse}
Baoxiong Jia, Yixin Chen, Huangyue Yu, Yan Wang, Xuesong Niu, Tengyu Liu, Qing Li, and Siyuan Huang.
\newblock Sceneverse: Scaling 3d vision-language learning for grounded scene understanding.
\newblock In \emph{ECCV}, 2024.

\bibitem[Jiang et~al.(2020)Jiang, Zhao, Shi, Liu, Fu, and Jia]{Pointgroup}
Li Jiang, Hengshuang Zhao, Shaoshuai Shi, Shu Liu, Chi-Wing Fu, and Jiaya Jia.
\newblock Pointgroup: Dual-set point grouping for 3d instance segmentation.
\newblock In \emph{CVPR}, 2020.

\bibitem[Jin et~al.(2023)Jin, Hayat, Yang, Guo, and Lei]{3dvlp}
Zhao Jin, Munawar Hayat, Yuwei Yang, Yulan Guo, and Yinjie Lei.
\newblock Context-aware alignment and mutual masking for 3d-language pre-training.
\newblock In \emph{CVPR}, 2023.

\bibitem[Kang et~al.(2024{\natexlab{a}})Kang, Huang, Shang, Shah, and Yan]{Robin3D}
Weitai Kang, Haifeng Huang, Yuzhang Shang, Mubarak Shah, and Yan Yan.
\newblock Robin3d: Improving 3d large language model via robust instruction tuning.
\newblock \emph{arXiv:2410.00255}, 2024{\natexlab{a}}.

\bibitem[Kang et~al.(2024{\natexlab{b}})Kang, Qu, Kini, Wei, Shah, and Yan]{Intent3D}
Weitai Kang, Mengxue Qu, Jyoti Kini, Yunchao Wei, Mubarak Shah, and Yan Yan.
\newblock Intent3d: 3d object detection in rgb-d scans based on human intention.
\newblock \emph{arXiv:2405.18295}, 2024{\natexlab{b}}.

\bibitem[Khanna et~al.(2024)Khanna, Mao, Jiang, Haresh, Shacklett, Batra, Clegg, Undersander, Chang, and Savva]{hssd-200}
Mukul Khanna, Yongsen Mao, Hanxiao Jiang, Sanjay Haresh, Brennan Shacklett, Dhruv Batra, Alexander Clegg, Eric Undersander, Angel~X Chang, and Manolis Savva.
\newblock Habitat synthetic scenes dataset (hssd-200): An analysis of 3d scene scale and realism tradeoffs for objectgoal navigation.
\newblock In \emph{CVPR}, 2024.

\bibitem[Lai et~al.(2024)Lai, Tian, Chen, Li, Yuan, Liu, and Jia]{Lisa}
Xin Lai, Zhuotao Tian, Yukang Chen, Yanwei Li, Yuhui Yuan, Shu Liu, and Jiaya Jia.
\newblock Lisa: Reasoning segmentation via large language model.
\newblock In \emph{CVPR}, 2024.

\bibitem[Li et~al.(2024)Li, Zhang, Guo, Zhang, Li, Zhang, Zhang, Li, Liu, and Li]{LLaVA-OneVision}
Bo Li, Yuanhan Zhang, Dong Guo, Renrui Zhang, Feng Li, Hao Zhang, Kaichen Zhang, Yanwei Li, Ziwei Liu, and Chunyuan Li.
\newblock Llava-onevision: Easy visual task transfer.
\newblock \emph{arXiv:2408.03326}, 2024.

\bibitem[Li et~al.(2025)Li, Wu, Zhang, Xia, Mao, Dong, Vulić, and Wei]{MVoT}
Chengzu Li, Wenshan Wu, Huanyu Zhang, Yan Xia, Shaoguang Mao, Li Dong, Ivan Vulić, and Furu Wei.
\newblock Imagine while reasoning in space: Multimodal visualization-of-thought.
\newblock \emph{arXiv:2501.07542}, 2025.

\bibitem[Li et~al.(2022)Li, Li, Xiong, and Hoi]{Blip}
Junnan Li, Dongxu Li, Caiming Xiong, and Steven Hoi.
\newblock Blip: Bootstrapping language-image pre-training for unified vision-language understanding and generation.
\newblock In \emph{ICML}, 2022.

\bibitem[Li et~al.(2023{\natexlab{a}})Li, Li, Savarese, and Hoi]{blip2}
Junnan Li, Dongxu Li, Silvio Savarese, and Steven Hoi.
\newblock Blip-2: Bootstrapping language-image pre-training with frozen image encoders and large language models.
\newblock \emph{arXiv:2301.12597}, 2023{\natexlab{a}}.

\bibitem[Li et~al.(2023{\natexlab{b}})Li, Wang, He, Li, Wang, Liu, Wang, Xu, Chen, Luo, Wang, and Qiao]{MVBench}
Kunchang Li, Yali Wang, Yinan He, Yizhuo Li, Yi Wang, Yi Liu, Zun Wang, Jilan Xu, Guo Chen, Ping Luo, Limin Wang, and Yu Qiao.
\newblock Mvbench: A comprehensive multi-modal video understanding benchmark.
\newblock \emph{arXiv:2311.17005}, 2023{\natexlab{b}}.

\bibitem[Li et~al.(2023{\natexlab{c}})Li, Wang, and Jia]{LLaMA-VID}
Yanwei Li, Chengyao Wang, and Jiaya Jia.
\newblock Llama-vid: An image is worth 2 tokens in large language models.
\newblock \emph{arXiv:2311.17043}, 2023{\natexlab{c}}.

\bibitem[Liu et~al.(2024{\natexlab{a}})Liu, Dong, Wang, Rao, Tang, Ma, and Krishna]{cc}
Benlin Liu, Yuhao Dong, Yiqin Wang, Yongming Rao, Yansong Tang, Wei-Chiu Ma, and Ranjay Krishna.
\newblock Coarse correspondence elicit 3d spacetime understanding in multimodal language model.
\newblock \emph{arXiv:2408.00754}, 2024{\natexlab{a}}.

\bibitem[Liu et~al.(2023)Liu, Li, Wu, and Lee]{llava}
Haotian Liu, Chunyuan Li, Qingyang Wu, and Yong~Jae Lee.
\newblock Visual instruction tuning.
\newblock In \emph{NeurIPS}, 2023.

\bibitem[Liu et~al.(2024{\natexlab{b}})Liu, Li, Li, and Lee]{llava1.5}
Haotian Liu, Chunyuan Li, Yuheng Li, and Yong~Jae Lee.
\newblock Improved baselines with visual instruction tuning.
\newblock In \emph{CVPR}, 2024{\natexlab{b}}.

\bibitem[Liu et~al.(2024{\natexlab{c}})Liu, Li, Li, Li, Zhang, Shen, and Lee]{LLaVA-NeXT}
Haotian Liu, Chunyuan Li, Yuheng Li, Bo Li, Yuanhan Zhang, Sheng Shen, and Yong~Jae Lee.
\newblock Llava-next: Improved reasoning, ocr, and world knowledge, 2024{\natexlab{c}}.

\bibitem[Liu et~al.(2024{\natexlab{d}})Liu, Dong, Wang, Yang, Fan, and Chen]{SLAM3R}
Yuzheng Liu, Siyan Dong, Shuzhe Wang, Yanchao Yang, Qingnan Fan, and Baoquan Chen.
\newblock Slam3r: Real-time dense scene reconstruction from monocular rgb videos.
\newblock \emph{arXiv:2412.09401}, 2024{\natexlab{d}}.

\bibitem[Luo et~al.(2022)Luo, Fu, Kong, Gao, Ren, Shen, Xia, and Liu]{3D-SPS}
Junyu Luo, Jiahui Fu, Xianghao Kong, Chen Gao, Haibing Ren, Hao Shen, Huaxia Xia, and Si Liu.
\newblock 3d-sps: Single-stage 3d visual grounding via referred point progressive selection.
\newblock In \emph{CVPR}, 2022.

\bibitem[Luo et~al.(2023)Luo, Zhao, Yang, Dong, Qiu, Lu, Wang, and Wei]{Valley}
Ruipu Luo, Ziwang Zhao, Min Yang, Junwei Dong, Minghui Qiu, Pengcheng Lu, Tao Wang, and Zhongyu Wei.
\newblock Valley: Video assistant with large language model enhanced ability.
\newblock \emph{arXiv:2306.07207}, 2023.

\bibitem[Ma et~al.(2023)Ma, Yong, Zheng, Li, Liang, Zhu, and Huang]{sqa3d}
Xiaojian Ma, Silong Yong, Zilong Zheng, Qing Li, Yitao Liang, Song-Chun Zhu, and Siyuan Huang.
\newblock Sqa3d: Situated question answering in 3d scenes.
\newblock In \emph{ICLR}, 2023.

\bibitem[Maaz et~al.(2023)Maaz, Rasheed, Khan, and Khan]{Video-chatgpt}
Muhammad Maaz, Hanoona Rasheed, Salman Khan, and Fahad~Shahbaz Khan.
\newblock Video-chatgpt: Towards detailed video understanding via large vision and language models.
\newblock \emph{arXiv:2306.05424}, 2023.

\bibitem[Man et~al.(2024)Man, Zheng, Bao, Hebert, Gui, and Wang]{Lexicon3D}
Yunze Man, Shuhong Zheng, Zhipeng Bao, Martial Hebert, Liang-Yan Gui, and Yu-Xiong Wang.
\newblock Lexicon3d: Probing visual foundation models for complex 3d scene understanding.
\newblock In \emph{NeurIPS}, 2024.

\bibitem[Mao et~al.(2022)Mao, Zhang, Jiang, Chang, and Savva]{MultiScan}
Yongsen Mao, Yiming Zhang, Hanxiao Jiang, Angel Chang, and Manolis Savva.
\newblock Multiscan: Scalable rgbd scanning for 3d environments with articulated objects.
\newblock In \emph{NeurIPS}, 2022.

\bibitem[Misra et~al.(2021)Misra, Girdhar, and Joulin]{3DETR}
Ishan Misra, Rohit Girdhar, and Armand Joulin.
\newblock An end-to-end transformer model for 3d object detection.
\newblock In \emph{ICCV}, 2021.

\bibitem[OpenAI(2023)]{gpt4}
OpenAI.
\newblock Gpt-4 technical report.
\newblock \emph{arXiv:2303.08774}, 2023.

\bibitem[{OpenAI}(2024)]{gpt4o}
{OpenAI}.
\newblock Gpt-4o: A large language model by openai.
\newblock \url{https://openai.com/research/gpt-4o}, 2024.

\bibitem[Peng et~al.(2023{\natexlab{a}})Peng, Genova, Jiang, Tagliasacchi, Pollefeys, Funkhouser, et~al.]{Openscene}
Songyou Peng, Kyle Genova, Chiyu Jiang, Andrea Tagliasacchi, Marc Pollefeys, Thomas Funkhouser, et~al.
\newblock Openscene: 3d scene understanding with open vocabularies.
\newblock In \emph{CVPR}, 2023{\natexlab{a}}.

\bibitem[Peng et~al.(2023{\natexlab{b}})Peng, Wang, Dong, Hao, Huang, Ma, and Wei]{kosmos2}
Zhiliang Peng, Wenhui Wang, Li Dong, Yaru Hao, Shaohan Huang, Shuming Ma, and Furu Wei.
\newblock Kosmos-2: Grounding multimodal large language models to the world.
\newblock \emph{arXiv:2306.14824}, 2023{\natexlab{b}}.

\bibitem[Qi et~al.(2019)Qi, Litany, He, and Guibas]{VoteNet}
Charles~R Qi, Or Litany, Kaiming He, and Leonidas~J Guibas.
\newblock Deep hough voting for 3d object detection in point clouds.
\newblock In \emph{ICCV}, 2019.

\bibitem[Qi et~al.(2024{\natexlab{a}})Qi, Dong, Zhang, Geng, Han, Ge, Yi, and Ma]{shapellm}
Zekun Qi, Runpei Dong, Shaochen Zhang, Haoran Geng, Chunrui Han, Zheng Ge, Li Yi, and Kaisheng Ma.
\newblock Shapellm: Universal 3d object understanding for embodied interaction.
\newblock In \emph{ECCV}, 2024{\natexlab{a}}.

\bibitem[Qi et~al.(2024{\natexlab{b}})Qi, Fang, Sun, Wu, Wu, Wang, Lin, and Zhao]{GPT4Point}
Zhangyang Qi, Ye Fang, Zeyi Sun, Xiaoyang Wu, Tong Wu, Jiaqi Wang, Dahua Lin, and Hengshuang Zhao.
\newblock Gpt4point: A unified framework for point-language understanding and generation.
\newblock In \emph{CVPR}, 2024{\natexlab{b}}.

\bibitem[Radford et~al.(2021)Radford, Kim, Hallacy, Ramesh, Goh, Agarwal, Sastry, Askell, Mishkin, Clark, et~al.]{CLIP}
Alec Radford, Jong~Wook Kim, Chris Hallacy, Aditya Ramesh, Gabriel Goh, Sandhini Agarwal, Girish Sastry, Amanda Askell, Pamela Mishkin, Jack Clark, et~al.
\newblock Learning transferable visual models from natural language supervision.
\newblock In \emph{ICML}, 2021.

\bibitem[Ramakrishnan et~al.(2021)Ramakrishnan, Gokaslan, Wijmans, Maksymets, Clegg, Turner, Undersander, Galuba, Westbury, Chang, et~al.]{HM3D}
Santhosh~K Ramakrishnan, Aaron Gokaslan, Erik Wijmans, Oleksandr Maksymets, Alex Clegg, John Turner, Eric Undersander, Wojciech Galuba, Andrew Westbury, Angel~X Chang, et~al.
\newblock Habitat-matterport 3d dataset (hm3d): 1000 large-scale 3d environments for embodied ai.
\newblock In \emph{NeurIPS}, 2021.

\bibitem[Rasheed et~al.(2023)Rasheed, Maaz, Shaji, Shaker, Khan, Cholakkal, Anwer, Xing, Yang, and Khan]{Glamm}
Hanoona Rasheed, Muhammad Maaz, Sahal Shaji, Abdelrahman Shaker, Salman Khan, Hisham Cholakkal, Rao~M Anwer, Erix Xing, Ming-Hsuan Yang, and Fahad~S Khan.
\newblock Glamm: Pixel grounding large multimodal model.
\newblock \emph{arXiv:2311.03356}, 2023.

\bibitem[Ren et~al.(2023)Ren, Yao, Li, Sun, and Hou]{TimeChat}
Shuhuai Ren, Linli Yao, Shicheng Li, Xu Sun, and Lu Hou.
\newblock Timechat: A time-sensitive multimodal large language model for long video understanding.
\newblock \emph{arXiv:2312.02051}, 2023.

\bibitem[Schuhmann et~al.(2022)Schuhmann, Beaumont, Vencu, Gordon, Wightman, Cherti, Coombes, Katta, Mullis, Wortsman, et~al.]{Laion-5b}
Christoph Schuhmann, Romain Beaumont, Richard Vencu, Cade Gordon, Ross Wightman, Mehdi Cherti, Theo Coombes, Aarush Katta, Clayton Mullis, Mitchell Wortsman, et~al.
\newblock Laion-5b: An open large-scale dataset for training next generation image-text models.
\newblock In \emph{NeurIPS}, 2022.

\bibitem[Schult et~al.(2023)Schult, Engelmann, Hermans, Litany, Tang, and Leibe]{Mask3D}
Jonas Schult, Francis Engelmann, Alexander Hermans, Or Litany, Siyu Tang, and Bastian Leibe.
\newblock Mask3d: Mask transformer for 3d semantic instance segmentation.
\newblock In \emph{ICRA}, 2023.

\bibitem[Song et~al.(2023)Song, Chai, Wang, Zhang, Zhou, Wu, Guo, Ye, Lu, Hwang, et~al.]{Moviechat}
Enxin Song, Wenhao Chai, Guanhong Wang, Yucheng Zhang, Haoyang Zhou, Feiyang Wu, Xun Guo, Tian Ye, Yan Lu, Jenq-Neng Hwang, et~al.
\newblock Moviechat: From dense token to sparse memory for long video understanding.
\newblock \emph{arXiv:2307.16449}, 2023.

\bibitem[Straub et~al.(2019)Straub, Whelan, Ma, Chen, Wijmans, Green, Engel, Mur-Artal, Ren, Verma, et~al.]{Replica}
Julian Straub, Thomas Whelan, Lingni Ma, Yufan Chen, Erik Wijmans, Simon Green, Jakob~J Engel, Raul Mur-Artal, Carl Ren, Shobhit Verma, et~al.
\newblock The replica dataset: A digital replica of indoor spaces.
\newblock \emph{arXiv:1906.05797}, 2019.

\bibitem[Sun et~al.(2023)Sun, Yu, Cui, Zhang, Zhang, Wang, Gao, Liu, Huang, and Wang]{Emu}
Quan Sun, Qiying Yu, Yufeng Cui, Fan Zhang, Xiaosong Zhang, Yueze Wang, Hongcheng Gao, Jingjing Liu, Tiejun Huang, and Xinlong Wang.
\newblock Emu: Generative pretraining in multimodality.
\newblock In \emph{ICLR}, 2023.

\bibitem[Takmaz et~al.(2023)Takmaz, Fedele, Sumner, Pollefeys, Tombari, and Engelmann]{OpenMask3D}
Ay{\c{c}}a Takmaz, Elisabetta Fedele, Robert~W Sumner, Marc Pollefeys, Federico Tombari, and Francis Engelmann.
\newblock Openmask3d: Open-vocabulary 3d instance segmentation.
\newblock In \emph{NeurIPS}, 2023.

\bibitem[Team(2024{\natexlab{a}})]{Gemini}
Gemini Team.
\newblock Gemini 1.5: Unlocking multimodal understanding across millions of tokens of context.
\newblock \emph{arXiv:2403.05530}, 2024{\natexlab{a}}.

\bibitem[Team and DeepMind(2024)]{gemma2}
Gemma Team and Google DeepMind.
\newblock Gemma 2: Improving open language models at a practical size.
\newblock \emph{arXiv:2408.00118}, 2024.

\bibitem[Team(2024{\natexlab{b}})]{internlm2}
InternLM Team.
\newblock Internlm2 technical report.
\newblock \emph{arXiv:2403.17297}, 2024{\natexlab{b}}.

\bibitem[Team(2024{\natexlab{c}})]{llama3}
Llama Team.
\newblock The llama 3 herd of models.
\newblock \emph{arXiv:2407.21783}, 2024{\natexlab{c}}.

\bibitem[Team(2024{\natexlab{d}})]{qwen25}
Qwen Team.
\newblock Qwen2.5 technical report.
\newblock \emph{arXiv:2412.15115}, 2024{\natexlab{d}}.

\bibitem[Unal et~al.(2024)Unal, Sakaridis, Saha, and Gool]{ConcreteNet}
Ozan Unal, Christos Sakaridis, Suman Saha, and Luc~Van Gool.
\newblock Four ways to improve verbo-visual fusion for dense 3d visual grounding.
\newblock In \emph{ECCV}, 2024.

\bibitem[Vu et~al.(2022)Vu, Kim, Luu, Nguyen, and Yoo]{SoftGroup}
Thang Vu, Kookhoi Kim, Tung~M Luu, Thanh Nguyen, and Chang~D Yoo.
\newblock Softgroup for 3d instance segmentation on point clouds.
\newblock In \emph{CVPR}, 2022.

\bibitem[Wald et~al.(2019)Wald, Avetisyan, Navab, Tombari, and Nie{\ss}ner]{Rio}
Johanna Wald, Armen Avetisyan, Nassir Navab, Federico Tombari, and Matthias Nie{\ss}ner.
\newblock Rio: 3d object instance re-localization in changing indoor environments.
\newblock In \emph{ICCV}, 2019.

\bibitem[Wang et~al.(2024{\natexlab{a}})Wang, Bai, Tan, Wang, Fan, Bai, Chen, Liu, Wang, Ge, Fan, Dang, Du, Ren, Men, Liu, Zhou, Zhou, and Lin]{Qwen2VL}
Peng Wang, Shuai Bai, Sinan Tan, Shijie Wang, Zhihao Fan, Jinze Bai, Keqin Chen, Xuejing Liu, Jialin Wang, Wenbin Ge, Yang Fan, Kai Dang, Mengfei Du, Xuancheng Ren, Rui Men, Dayiheng Liu, Chang Zhou, Jingren Zhou, and Junyang Lin.
\newblock Qwen2-vl: Enhancing vision-language model's perception of the world at any resolution.
\newblock \emph{arXiv:2409.12191}, 2024{\natexlab{a}}.

\bibitem[Wang et~al.(2024{\natexlab{b}})Wang, Mao, Zhu, Xu, Lyu, Li, Chen, Zhang, Chen, Xue, Liu, Lu, Lin, and Pang]{EmbodiedScan}
Tai Wang, Xiaohan Mao, Chenming Zhu, Runsen Xu, Ruiyuan Lyu, Peisen Li, Xiao Chen, Wenwei Zhang, Kai Chen, Tianfan Xue, Xihui Liu, Cewu Lu, Dahua Lin, and Jiangmiao Pang.
\newblock Embodiedscan: A holistic multi-modal 3d perception suite towards embodied ai.
\newblock In \emph{CVPR}, 2024{\natexlab{b}}.

\bibitem[Wang et~al.(2023{\natexlab{a}})Wang, Lv, Yu, Hong, Qi, Wang, Ji, Yang, Zhao, Song, et~al.]{cogvlm}
Weihan Wang, Qingsong Lv, Wenmeng Yu, Wenyi Hong, Ji Qi, Yan Wang, Junhui Ji, Zhuoyi Yang, Lei Zhao, Xixuan Song, et~al.
\newblock Cogvlm: Visual expert for pretrained language models.
\newblock \emph{arXiv:2311.03079}, 2023{\natexlab{a}}.

\bibitem[Wang et~al.(2023{\natexlab{b}})Wang, Huang, Zhao, Li, Cheng, Zhu, Yin, and Zhao]{3DRP-Net}
Zehan Wang, Haifeng Huang, Yang Zhao, Linjun Li, Xize Cheng, Yichen Zhu, Aoxiong Yin, and Zhou Zhao.
\newblock 3drp-net: 3d relative position-aware network for 3d visual grounding.
\newblock In \emph{EMNLP}, 2023{\natexlab{b}}.

\bibitem[Wang et~al.(2023{\natexlab{c}})Wang, Huang, Zhao, Zhang, and Zhao]{Chat-3D}
Zehan Wang, Haifeng Huang, Yang Zhao, Ziang Zhang, and Zhou Zhao.
\newblock Chat-3d: Data-efficiently tuning large language model for universal dialogue of 3d scenes.
\newblock \emph{arXiv:2308.08769}, 2023{\natexlab{c}}.

\bibitem[Wu et~al.(2024{\natexlab{a}})Wu, Huang, and Wang]{DORA}
Ting-Yao Wu, Shih-Yang Huang, and Yung-Chen~Francis Wang.
\newblock Dora: 3d visual grounding with order-aware referring.
\newblock \emph{arXiv:2403.16539}, 2024{\natexlab{a}}.

\bibitem[Wu et~al.(2024{\natexlab{b}})Wu, Jiang, Wang, Liu, Liu, Qiao, Ouyang, He, and Zhao]{PTV3}
Xiaoyang Wu, Li Jiang, Peng-Shuai Wang, Zhijian Liu, Xihui Liu, Yu Qiao, Wanli Ouyang, Tong He, and Hengshuang Zhao.
\newblock Point transformer v3: Simpler, faster, stronger.
\newblock In \emph{CVPR}, 2024{\natexlab{b}}.

\bibitem[Wu et~al.(2023)Wu, Cheng, Zhang, Cheng, and Zhang]{EDA}
Yanmin Wu, Xinhua Cheng, Renrui Zhang, Zesen Cheng, and Jian Zhang.
\newblock Eda: Explicit text-decoupling and dense alignment for 3d visual grounding.
\newblock In \emph{CVPR}, 2023.

\bibitem[xAI(2024)]{RealWorldQA}
xAI, 2024.

\bibitem[Xu et~al.(2024{\natexlab{a}})Xu, Huang, Wang, Chen, Pang, and Lin]{VLM-Grounder}
Runsen Xu, Zhiwei Huang, Tai Wang, Yilun Chen, Jiangmiao Pang, and Dahua Lin.
\newblock Vlm-grounder: A vlm agent for zero-shot 3d visual grounding.
\newblock In \emph{CoRL}, 2024{\natexlab{a}}.

\bibitem[Xu et~al.(2024{\natexlab{b}})Xu, Wang, Wang, Chen, Pang, and Lin]{pointllm}
Runsen Xu, Xiaolong Wang, Tai Wang, Yilun Chen, Jiangmiao Pang, and Dahua Lin.
\newblock Pointllm: Empowering large language models to understand point clouds.
\newblock In \emph{ECCV}, 2024{\natexlab{b}}.

\bibitem[Xue et~al.(2023)Xue, Gao, Xing, Mart{\'\i}n-Mart{\'\i}n, Wu, Xiong, Xu, Niebles, and Savarese]{ULIP}
Le Xue, Mingfei Gao, Chen Xing, Roberto Mart{\'\i}n-Mart{\'\i}n, Jiajun Wu, Caiming Xiong, Ran Xu, Juan~Carlos Niebles, and Silvio Savarese.
\newblock Ulip: Learning a unified representation of language, images, and point clouds for 3d understanding.
\newblock In \emph{CVPR}, 2023.

\bibitem[Yang et~al.(2023)Yang, Ding, Wang, and Qi]{Regionplc}
Jihan Yang, Runyu Ding, Zhe Wang, and Xiaojuan Qi.
\newblock Regionplc: Regional point-language contrastive learning for open-world 3d scene understanding.
\newblock \emph{arXiv:2304.00962}, 2023.

\bibitem[Yang et~al.(2025)Yang, Yang, Gupta, Han, Fei-Fei, and Xie]{vsibench}
Jihan Yang, Shusheng Yang, Anjali Gupta, Rilyn Han, Li Fei-Fei, and Saining Xie.
\newblock {Thinking in Space: How Multimodal Large Language Models See, Remember and Recall Spaces}.
\newblock In \emph{CVPR}, 2025.

\bibitem[Yang et~al.(2021)Yang, Zhang, Wang, and Luo]{Sat}
Zhengyuan Yang, Songyang Zhang, Liwei Wang, and Jiebo Luo.
\newblock Sat: 2d semantics assisted training for 3d visual grounding.
\newblock In \emph{ICCV}, 2021.

\bibitem[Yao et~al.(2024)Yao, Yu, Zhang, Wang, Cui, Zhu, Cai, Li, Zhao, He, et~al.]{MiniCPM-V}
Yuan Yao, Tianyu Yu, Ao Zhang, Chongyi Wang, Junbo Cui, Hongji Zhu, Tianchi Cai, Haoyu Li, Weilin Zhao, Zhihui He, et~al.
\newblock Minicpm-v: A gpt-4v level mllm on your phone.
\newblock \emph{arXiv:2408.01800}, 2024.

\bibitem[Ye et~al.(2022)Ye, Chen, Han, and Liao]{3dqa}
Shuquan Ye, Dongdong Chen, Songfang Han, and Jing Liao.
\newblock 3d question answering.
\newblock \emph{TVCG}, 2022.

\bibitem[Yuan et~al.(2021)Yuan, Yan, Liao, Zhang, Li, and Cui]{Instancerefer}
Zhihao Yuan, Xu Yan, Yinghong Liao, Ruimao Zhang, Zhen Li, and Shuguang Cui.
\newblock Instancerefer: Cooperative holistic understanding for visual grounding on point clouds through instance multi-level contextual referring.
\newblock In \emph{ICCV}, 2021.

\bibitem[Yuan et~al.(2022)Yuan, Yan, Liao, Guo, Li, Cui, and Li]{X-trans2cap}
Zhihao Yuan, Xu Yan, Yinghong Liao, Yao Guo, Guanbin Li, Shuguang Cui, and Zhen Li.
\newblock X-trans2cap: Cross-modal knowledge transfer using transformer for 3d dense captioning.
\newblock In \emph{CVPR}, 2022.

\bibitem[Yuan~Liu(2023)]{MMBench}
Yuanhan Zhang et.~al Yuan~Liu, Haodong~Duan.
\newblock Mmbench: Is your multi-modal model an all-around player?
\newblock \emph{arXiv:2307.06281}, 2023.

\bibitem[Yunhan~Yang(2023)]{SAM3D}
Xiaoyang Wu~et.al Yunhan~Yang.
\newblock Sam3d: Segment anything in 3d scenes.
\newblock In \emph{ICCV Workshop}, 2023.

\bibitem[Zhang et~al.(2023{\natexlab{a}})Zhang, Li, and Bing]{Video-llama}
Hang Zhang, Xin Li, and Lidong Bing.
\newblock Video-llama: An instruction-tuned audio-visual language model for video understanding.
\newblock \emph{arXiv:2306.02858}, 2023{\natexlab{a}}.

\bibitem[Zhang et~al.(2022)Zhang, Guo, Zhang, Li, Miao, Cui, Qiao, Gao, and Li]{Pointclip}
Renrui Zhang, Ziyu Guo, Wei Zhang, Kunchang Li, Xupeng Miao, Bin Cui, Yu Qiao, Peng Gao, and Hongsheng Li.
\newblock Pointclip: Point cloud understanding by clip.
\newblock In \emph{CVPR}, 2022.

\bibitem[Zhang et~al.(2023{\natexlab{b}})Zhang, Wang, Qiao, Gao, and Li]{I2P-MAE}
Renrui Zhang, Liuhui Wang, Yu Qiao, Peng Gao, and Hongsheng Li.
\newblock Learning 3d representations from 2d pre-trained models via image-to-point masked autoencoders.
\newblock In \emph{CVPR}, 2023{\natexlab{b}}.

\bibitem[Zhang et~al.(2024)Zhang, Huang, Deng, Tang, Ouyang, He, and Zhang]{Agent3D-Zero}
Sha Zhang, Di Huang, Jiajun Deng, Shixiang Tang, Wanli Ouyang, Tong He, and Yanyong Zhang.
\newblock Agent3d-zero: An agent for zero-shot 3d understanding.
\newblock In \emph{ECCV}, 2024.

\bibitem[Zhang et~al.(2023{\natexlab{c}})Zhang, Gong, and Chang]{Multi3DRefer}
Yiming Zhang, ZeMing Gong, and Angel~X Chang.
\newblock Multi3drefer: Grounding text description to multiple 3d objects.
\newblock In \emph{ICCV}, 2023{\natexlab{c}}.

\bibitem[Zhao et~al.(2021)Zhao, Cai, Sheng, and Xu]{3DVG-Transformer}
Lichen Zhao, Daigang Cai, Lu Sheng, and Dong Xu.
\newblock 3dvg-transformer: Relation modeling for visual grounding on point clouds.
\newblock In \emph{ICCV}, 2021.

\bibitem[Zheng et~al.(2025)Zheng, Huang, and Wang]{Video-3D-LLM}
Duo Zheng, Shijia Huang, and Liwei Wang.
\newblock Video-3d llm: Learning position-aware video representation for 3d scene understanding.
\newblock In \emph{CVPR}, 2025.

\bibitem[Zheng et~al.(2020)Zheng, Zhang, Li, Tang, Gao, and Zhou]{Structured3d}
Jia Zheng, Junfei Zhang, Jing Li, Rui Tang, Shenghua Gao, and Zihan Zhou.
\newblock Structured3d: A large photo-realistic dataset for structured 3d modeling.
\newblock In \emph{ECCV}, 2020.

\bibitem[Zhu et~al.(2024{\natexlab{a}})Zhu, Wang, Zhang, Pang, and Liu]{LLaVA-3D}
Chenming Zhu, Tai Wang, Wenwei Zhang, Jiangmiao Pang, and Xihui Liu.
\newblock Llava-3d: A simple yet effective pathway to empowering lmms with 3d-awareness.
\newblock \emph{arXiv:2409.18125}, 2024{\natexlab{a}}.

\bibitem[Zhu et~al.(2023{\natexlab{a}})Zhu, Chen, Shen, Li, and Elhoseiny]{minigpt4}
Deyao Zhu, Jun Chen, Xiaoqian Shen, Xiang Li, and Mohamed Elhoseiny.
\newblock Minigpt-4: Enhancing vision-language understanding with advanced large language models.
\newblock \emph{arXiv:2304.10592}, 2023{\natexlab{a}}.

\bibitem[Zhu et~al.(2023{\natexlab{b}})Zhu, Ma, Chen, Deng, Huang, and Li]{3D-VisTA}
Ziyu Zhu, Xiaojian Ma, Yixin Chen, Zhidong Deng, Siyuan Huang, and Qing Li.
\newblock 3d-vista: Pre-trained transformer for 3d vision and text alignment.
\newblock In \emph{ICCV}, 2023{\natexlab{b}}.

\bibitem[Zhu et~al.(2024{\natexlab{b}})Zhu, Zhang, Ma, Niu, Chen, Jia, Deng, Huang, and Li]{PQ3D}
Ziyu Zhu, Zhuofan Zhang, Xiaojian Ma, Xuesong Niu, Yixin Chen, Baoxiong Jia, Zhidong Deng, Siyuan Huang, and Qing Li.
\newblock Unifying 3d vision-language understanding via promptable queries.
\newblock In \emph{ECCV}, 2024{\natexlab{b}}.

\end{thebibliography}

}
\end{document}